\documentclass{article}

\usepackage{microtype}
\usepackage{graphicx}
\usepackage{booktabs} 
\usepackage{amsmath} 
\usepackage{amsfonts}
\usepackage{amssymb}

\usepackage{hyperref}
\usepackage{enumitem}
\usepackage[normalem]{ulem}
\usepackage[noend]{algorithmic}
\usepackage{multirow}
\usepackage{xspace}
\usepackage{subcaption}
\usepackage{wrapfig}

\newcommand{\approach}{\textsc{MACE}\xspace}

\usepackage[accepted]{mlsys2025}

\mlsystitlerunning{\approach: A Hybrid LLM Serving System with Colocated SLO-aware Continuous Retraining Alignment}
\begin{document}

\twocolumn[
\mlsystitle{\approach: A Hybrid LLM Serving System with Colocated SLO-aware Continuous Retraining Alignment}



\mlsyssetsymbol{equal}{*}

\begin{mlsysauthorlist}
\mlsysauthor{Yufei Li}{ucr}
\mlsysauthor{Yu Fu}{ucr}
\mlsysauthor{Yue Dong}{ucr}
\mlsysauthor{Cong Liu}{ucr}
\end{mlsysauthorlist}

\mlsysaffiliation{ucr}{University of California, Riverside}

\mlsyscorrespondingauthor{Yufei Li}{yli927@ucr.edu}
\mlsyscorrespondingauthor{Cong Liu}{congl@ucr.edu}

\mlsyskeywords{Machine Learning, MLSys}

\vskip 0.3in

\begin{abstract}
Large language models (LLMs) deployed on edge servers are increasingly used in latency-sensitive applications such as personalized assistants, recommendation, and content moderation. However, the non-stationary nature of user data necessitates frequent retraining, which introduces a fundamental tension between \emph{inference latency} and \emph{model accuracy} under constrained GPU resources. Existing retraining strategies either delay model updates, over-commit resources to retraining, or overlook iteration-level retraining granularity. In this paper, we identify that iteration-level scheduling is crucial for adapting retraining frequency to model drift without violating service-level objectives (SLOs). We propose \approach, a hybrid LLM system that \emph{colocates} concurrent inference (prefill, decode) and fine-tuning, with intelligent memory management to maximize task performance while promising inference throughput. 
\approach leverages the insight that not all model updates equally affect output alignment and allocates GPU cycles accordingly to balance throughput, latency, and update freshness.
Our trace-driven evaluation shows that \approach matches or exceeds continuous retraining while reducing inference latency by up to 63\% and maintaining throughput under resource constraints. Compared to periodic retraining, \approach improves latency breakdown across prefill, decode, and finetune stages, and sustains GPU utilization above 85\% in NVIDIA AGX Orin. These results demonstrate that iteration-level hybrid scheduling is a promising direction for deploying LLMs with continual learning capabilities on edge platforms.




\end{abstract}
]



\printAffiliationsAndNotice{}  

\section{Introduction}
\label{sec:intro}

Large language models (LLMs) have demonstrated impressive capabilities across diverse real-world applications, including open-domain question answering, code generation, and mathematical reasoning. 
As a result, LLMs have been widely adopted in interactive systems such as chat assistants~\cite{achiam2023gpt,qiao2024autoact}, search engines~\cite{perplexity_ai}, multimodal agents~\cite{lin2025creativity,sarukkai2025self}, and real-time code assistants~\cite{github_copilot}. 
These systems process user queries in real-time and have increasingly shifted toward edge deployment, where models are hosted on local edge servers closer to the user~\cite{fanton_edge_2021,shubha2023adainf}.


Edge servers offer low-latency responses and improved data privacy, complying with regional policies like General Data Protection Regulation (GDPR) that prohibit the transmission of sensitive data to distant cloud regions~\cite{bhardwaj2022ekya}. However, edge servers often have limited GPU capacity~\cite{mittal2021mu}, making it challenging to concurrently serve inference requests and support continual adaptation of models to user feedback or real-time data shifts~\cite{bhardwaj2022ekya}. For instance, user responses may signal changes in preference (e.g., political leaning, toxicity tolerance)~\cite{liang2025comprehensive}, requiring the system to fine-tune the model to reflect these updated objectives—departing from generalized alignment toward personalized alignment.

While compression methods (e.g., quantization, distillation) can reduce inference costs, their limited generalization capacity makes them highly sensitive to data distribution shifts~\cite{mittal2021mu,shubha2023adainf}. Therefore, continual retraining (e.g., every 30–60 seconds) becomes necessary to maintain alignment~\cite{bhardwaj2022ekya,shubha2023adainf}, using user-specific preferences and temporal contexts extracted from live inference interactions.
Unlike conventional supervised training pipelines, labels for LLM alignment data are not manually annotated. Instead, they are derived from implicit user feedback signals. A common approach is to use preference-based supervision, where users are presented with multiple generated responses and asked to select the most preferred one.
User feedback is converted into supervision through preference signals (e.g., chosen, rejected), enabling ranking-based pairwise losses like Direct Preference Optimization (DPO)~\cite{rafailov2023direct}. Additional implicit signals—e.g., thumbs-up/down, skip rate, or dwell time—can be mapped to preferences or reward scores. In multi-turn dialogs, user rephrasing or corrections also serve as supervision. Over time, this generates a personalized feedback corpus for fine-tuning models toward user-specific goals, without centralized fine-tuning or privacy violations.

\begin{figure}
    \centering
    \includegraphics[width=\linewidth]{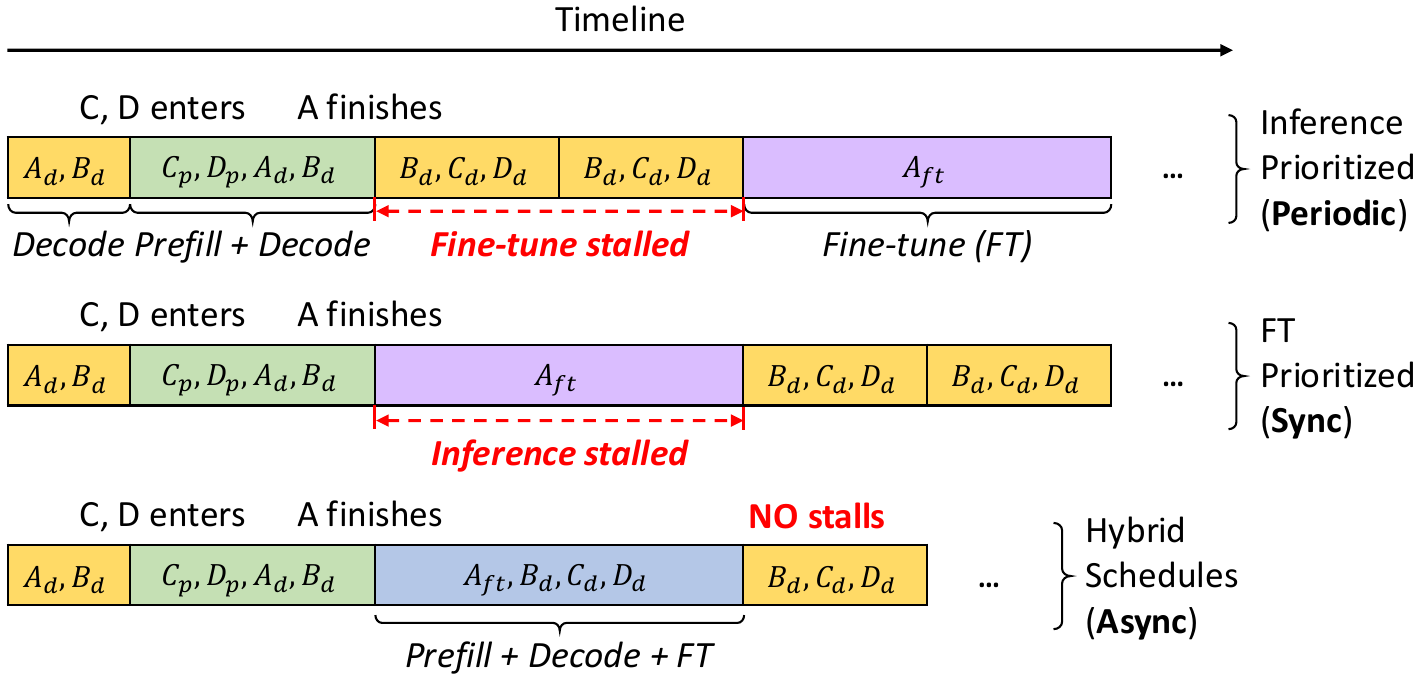}
    \caption{
    Requests A, B, C, D arrive over time. Subscripts $p$ and $d$ indicate prefill and decode iterations, while $ft$ marks fine-tuning. \textbf{Periodic} retraining delays model updates for A due to inference priority. \textbf{Sync} retraining preempts decodes for B, C, D. \textbf{Async} (hybrid) schedule \emph{colocates} A$_{ft}$, B$_{d}$, C$_{d}$, D$_{d}$ into the same iteration, reducing latency and ensuring B, C, D benefit (in subsequent iterations) from the updated model—without stalling either workload.
    }
    \label{fig:schedules}
    \vspace{-5pt}
\end{figure}

However, co-running fine-tuning and inference on a resource-constrained GPU introduces a fundamental scheduling dilemma. Figure~\ref{fig:schedules} illustrates the trade-offs among three commonly used scheduling strategies, building upon iteration-based continuous batching~\cite{yu2022orca}, where time is divided into discrete iterations, and each iteration forms a batch of tasks for the model to process concurrently:
\begin{itemize}[nosep, leftmargin=*]
    \item \emph{Periodic} retraining defers model updates to fixed intervals, preserving inference responsiveness. Yet, this can delay alignment—decoding requests (B, C, D) are served using outdated models, reducing accuracy.
    \item \emph{Continuous} (sync) retraining immediately preempts inference to perform fine-tuning, ensuring alignment but often stalling inference requests and causing SLO violations.
    \item \emph{Hybrid} (async) retraining aims to interleave both workloads within the same GPU iteration, packing complementary workloads to reduce idle fragments and maximize both alignment and inference throughput.
\end{itemize}
The core challenge arises from their conflicting resource usage patterns under tight GPU budgets. Inference—especially LLM autoregressive decoding—requires sequential generation and sustained attention memory, which makes batching difficult and memory-intensive~\cite{kwon2023efficient,agrawal2024taming}. Fine-tuning, on the other hand, demands high GPU memory bandwidth and model parameter updates, which can stall ongoing inference~\cite{choi2023envpipe}. 

To mitigate this tension, recent approaches employ \emph{parameter-efficient fine-tuning} (PEFT) methods such as low-rank adaptation (LoRA)~\cite{hu2022lora}, which injects trainable low-rank matrices into the frozen weights of the base model. PEFT significantly reduces the number of trainable parameters and memory overhead during training, making it a natural fit for resource-constrained edge environments. It allows fine-tuning to be performed with minimal disruption to inference serving.
However, even with PEFT, co-executing fine-tuning and inference remains non-trivial. For example, LoRA introduces additional memory (for adapters and optimizer state), and its backward pass can conflict with autoregressive decoding, particularly when multiple user requests are queued. Moreover, \emph{resource fragmentation}—caused by variable-length inputs and dynamic arrival rates—can lead to suboptimal utilization and SLO violations (for inference requests) if not carefully managed~\cite{sun2024llumnix}.

To this end, we propose \approach, a hybrid execution framework that co-optimizes fine-tuning and inference on shared edge GPU resources. \approach introduces a fine-grained iteration-level hybrid scheduler that selects and packs both workloads together based on alignment potential, memory feasibility, and latency constraints. 
It adopts a \emph{unified scheduling} strategy that handles inference and continual learning under a shared memory and execution budget. 
Compared to traditional periodic or sync baselines, \approach adapts dynamically to the queue state and application demand, maximizing GPU utilization while minimizing latency for heterogeneous workloads. \approach is built on two key components:
\begin{itemize}[nosep, leftmargin=*]
    \item \emph{Memory-aware hybrid scheduler}, which jointly allocates prefill, decode, and fine-tune workloads per iteration using memory and alignment heuristics (\S\ref{sec:bin_packing}).

    \item \emph{Cache manager}, including prefix sharing at prefilling, enabling reuse of previously computed KV cache across requests to reduce response time and free memory, and KV cache pruning at decoding, which identifies and removes redundant attention heads or layers with negligible contribution to final output (\S\ref{sec:cache_management}).

\end{itemize}

We conduct extensive experiments on real-world traces from personalized chat datasets using Mistral-7B and LLaMA3-8B. As shown in Figures~\ref{fig:main_accuracy}, \ref{fig:main_throughput}, \approach consistently achieves the best trade-off between alignment accuracy and inference throughput, outperforming both periodic and synchronous baselines across retraining frequencies.

\section{Background}

Recent surge in deploying LLMs for real-time user interaction~\cite{shen2024large} has necessitated effective strategies for handling concurrent training and inference tasks. 
These workloads span responding to user queries and retraining aimed at aligning LLMs with human preferences~\cite{askell2021general} to ensure helpfulness~\cite{ethayarajh22a,li2025dr} and harmlessness~\cite{fu2024safety}.

\noindent\textbf{Alignment fine-tuning and personalization.}
Modern LLMs often rely on post-training alignment techniques, such as \emph{reinforcement learning from human feedback} (RLHF)~\cite{ouyang2022training} or DPO~\cite{rafailov2023direct} to better align generated responses with human preferences. To reduce the cost of full fine-tuning, PEFT like LoRA~\cite{hu2022lora} and surgical fine-tuning~\cite{lee2023surgical} have been widely adopted. 
LoRA introduces learnable low-rank matrices into attention layers, enabling efficient adaptation to new data without updating the backbone model. Surgical fine-tuning, on the other hand, identifies and updates only specific components of the model that are most impactful for alignment~\cite{shi2024understanding}. 
Such resource-efficient techniques are particularly advantageous for user personalization or continual alignment during deployment, where rapid adaptation to new preferences is essential.

\noindent\textbf{Advanced LLM serving systems.}
Deploying LLMs in real-time and high-throughput environments requires serving systems that can simultaneously satisfy low latency, high efficiency, and hardware resource constraints. Continuous batching has become the de facto design in modern LLM systems to handle dynamic request arrival, which group them on-the-fly to improve GPU utilization without introducing excessive queuing delays~\cite{li2023rt}, as shown in Algorithm~\ref{alg:continuous_batching} in Appendix~\ref{app:method}. Systems like Orca~\cite{yu2022orca}, vLLM~\cite{kwon2023efficient}, Llumnix~\cite{sun2024llumnix}, and DistServe~\cite{zhong2024distserve} have proposed various optimizations in this space. For example, Orca enables mixed prefill and decode execution via iteration (token)-level scheduling, and Llumnix proposes chunked prefill to pipeline long-context inputs. Despite implementation differences, these systems share a core challenge: \emph{How to efficiently pack heterogeneous requests with diverse memory and latency profiles?} We argue that continuous batching can be viewed through the lens of bin-scheduling—treating each GPU or sub-batch as a bin and scheduling tasks based on joint memory-latency fit. This abstraction forms the basis for our unified scheduling approach in later sections.

\section{Motivation Study}
\label{sec:motivation}

In this section, we first empirically show that continuous retraining outperforms periodic retraining in maintaining alignment quality over time. However, naïvely applying continuous retraining introduces new bottlenecks, as it interferes with inference workloads and degrades responsiveness. We further show that continuous retraining can be optimized via hybrid batching to achieve lower latency, enabling systems to deliver up-to-date and real-time aligned responses.

\subsection{Accuracy Benefits of Continuous Retraining}
\label{sec:alignment_metrics}

Traditional approaches to adapting LLMs to evolving user preferences often involve periodic retraining at fixed intervals, potentially leading to stale models that lag behind rapidly changing user needs. In contrast, continuous retraining allows models to adapt instantly, incorporating new information as it becomes available. 

\noindent\textbf{Alignment metrics.} To evaluate the effectiveness of continuous retraining in aligning LLMs with user preferences, we define the following setting. Let $\mathcal{D} = {(x, y^+, y^-)}$ denote a dataset of user preference annotations, where $x$ represents the input prompt, $y^+$ is the \emph{preferred} response, and $y^-$ is the \emph{dispreferred} response. We assess model $\pi_\theta$'s performance using the following alignment metrics:

\begin{itemize}[nosep, leftmargin=*]
    
    \item \textbf{Preference accuracy (win rate)}~\cite{zhang2025preference} $\uparrow$: Win rate measures the fraction of preference pairs where the model assigns a higher probability to the preferred response:
    \[
    \text{Win Rate} = \mathbb{E}_{(x, y^+, y^-) \sim \mathcal{D}} \left[ \mathbf{1}\left[\pi_\theta(y^+ \mid x) > \pi_\theta(y^- \mid x) \right] \right],
    \]
    where $\mathbf{1}[\cdot]$ is the indicator function. A higher win rate signifies better alignment with user preferences.
    
    \item \textbf{Contrastive log-probability difference (CLPD)}~\cite{chen2024noise} $\uparrow$: CLPD quantifies the average difference in log-probabilities between the preferred and less preferred responses:
    \[
    \text{CLPD} = \mathbb{E}_{(x, y^+, y^-) \sim \mathcal{D}} \left[ \log \pi_\theta(y^+ \mid x) - \log \pi_\theta(y^- \mid x) \right].
    \]
    A higher CLPD indicates a stronger preference for the preferred responses over the less preferred ones.
    
\end{itemize}

\noindent\textbf{Retraining setup.} To avoid the complexity of reinforcement learning, DPO~\cite{rafailov2023direct} provides a lightweight alternative by leveraging contrastive learning over user preferences. Instead of optimizing a reward model, DPO directly fine-tunes LLMs by comparing the preferred and dispreferred outputs, while regularizing against a reference policy $\pi_{\text{ref}}$ (e.g., the initial model). The goal is to maximize CLPD between $y^+$ and $y^-$, balanced by a regularization term that discourages excessive divergence from $\pi_{\text{ref}}$:
\[
\begin{gathered}
\mathcal{L}_{\text{DPO}} = \mathbb{E}_{(x, y^+, y^-) \sim \mathcal{D}} \left[
\log \sigma \left(
\beta \cdot 
\left( 
\Delta^+_\theta - \Delta^-_\theta 
\right) 
\right)
\right], \\
s.t. \; \Delta^+_\theta = \log \frac{\pi_\theta(y^+ \mid x)}{\pi_{\text{ref}}(y^+ \mid x)},\quad
\Delta^-_\theta = \log \frac{\pi_\theta(y^- \mid x)}{\pi_{\text{ref}}(y^- \mid x)}.
\end{gathered}
\]
Here, $\pi_\theta$ is the current model policy, $\beta$ is a temperature parameter controlling the sharpness of preference, and $\sigma(\cdot)$ denotes the sigmoid function. 

\begin{figure}[]
    \centering
    \includegraphics[width=\linewidth]{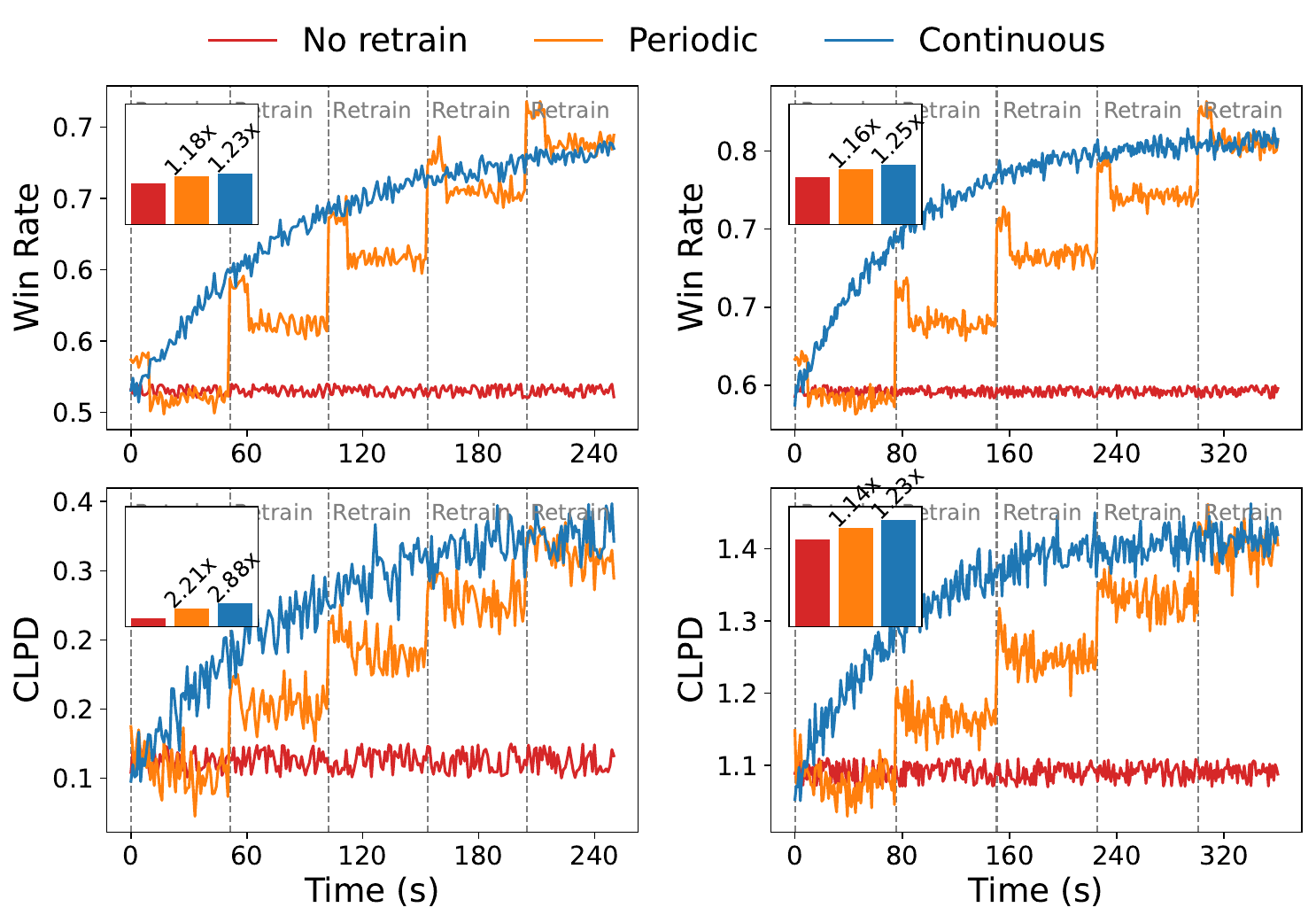}
    \caption{Win rate and CLPD \uline{over time} when serving Mistral-7B on \emph{Left}: RLHF and \emph{Right}: SHP dataset. The inset bar plots show the \uline{average metric value} of each method, highlighting the overall performance difference beyond temporal variations.}
    \label{fig:accuracy_comparison}
    \vspace{-10pt}
\end{figure}

\noindent\textbf{Empirical results.}
We study the impact of retraining strategies under a shared LLM serving system using two alignment fine-tuning benchmarks: Anthropic HH-RLHF~\cite{bai2022training} and StanfordNLP SHP~\cite{ethayarajh22a}. The former focuses on harmlessness, where preferred responses tend to reject unsafe or illegal requests, while the latter focuses on helpfulness, where preferred responses are more factually correct. 
Figure \ref{fig:accuracy_comparison} compares win rate and CLPD over time under different retraining strategies. We observe that:
\emph{Both periodic and continuous retraining consistently outperform the no-retrain baseline.} For instance, in the RLHF dataset (top-left), periodic retraining improves win rate by up to 1.19$\times$, while continuous retraining yields a 1.24$\times$ gain. Similar trends hold for CLPD, where the confidence gap grows even more substantially (bottom-left, 2.47$\times$ and 3.17$\times$, respectively).
\emph{Continuous retraining provides a larger performance boost.} Since retraining is triggered more frequently, continuous updates adapt the model faster to distributional shifts, leading to higher overall metrics compared to periodic retraining. This advantage is consistently seen across both datasets and both metrics.
\emph{CLPD reveals larger fluctuations and improvements than win rate.} While win rate rises more smoothly, CLPD exhibits stronger variance and larger relative gains. This is expected, as CLPD directly measures the preference confidence difference, which is more sensitive and interpretable in reflecting alignment strength.
\emph{Harmlessness alignment is harder than helpfulness alignment.} Comparing the two datasets, performance improvement is smaller for RLHF (left column) than for SHP (right column). This indicates that aligning models to avoid harmful behaviors is inherently more challenging than aligning them to be helpful.


\subsection{Optimization for Continuous Retraining}

\begin{figure}
    \centering 
    \includegraphics[width=\linewidth]{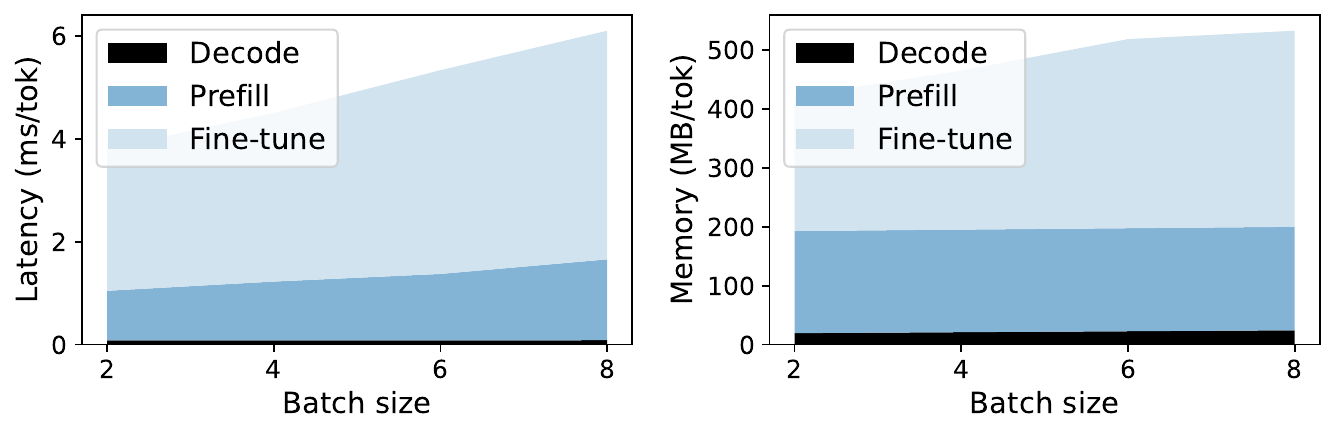}
    \caption{\emph{Left}: latency and \emph{Right}: memory per-token of three workloads for Mistral-7B on A6000 Ada across varying batch sizes.}
    \label{fig:latency_memory_profile}
    \vspace{-5pt}
\end{figure}

\noindent\textbf{Profiling  LLM workload heterogeneity.}
To effectively co-serve inference and continual fine-tuning for LLMs, it is essential to understand the latency and memory footprints of the constituent workloads. We profile the latency and memory usage of three major LLM operations—\textit{prefill}, \textit{decode}, and \textit{fine-tune}—across varying batch sizes, shown in Figure~\ref{fig:latency_memory_profile}.
We observe that:
\begin{itemize}[nosep,leftmargin=*]
    \item \emph{Prefill} latency and memory scale with input sequence length and batch size (linearly under FlashAttention~\cite{dao2022flashattention}), since each token attends to the full prefix.
    \item \emph{Decode} operations are much cheaper, as only one token is appended per iteration, but memory gradually increases due to KV cache accumulation~\cite{kwon2023efficient}.
    \item \emph{Fine-tune}, even with efficient PEFT methods like LoRA~\cite{hu2022lora}, exhibits significantly higher latency and memory consumption—making it the dominant bottleneck in edge inference-tuning pipelines. 
\end{itemize}

\begin{figure}[]
    \centering
    \includegraphics[width=\linewidth]{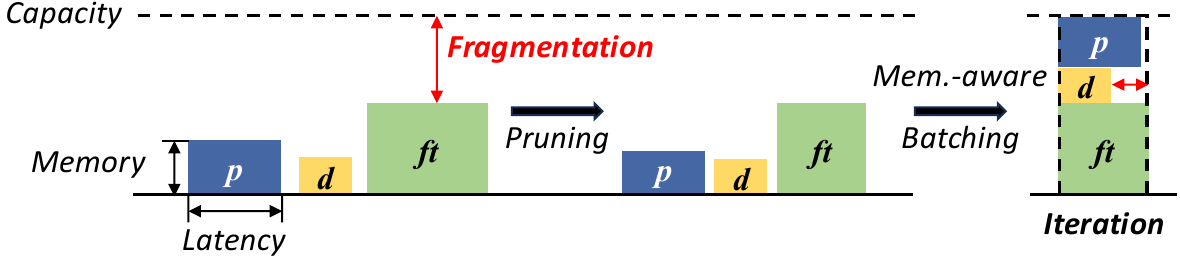}
    \caption{Abstracted memory–latency footprint for three workloads. Hybrid scheduling together with pruning techniques mitigates memory fragmentation and increases concurrency.}
    \label{fig:three_workloads}
    \vspace{-5pt}
\end{figure}

\noindent\textbf{Memory fragmentation and hybrid scheduling.}
Figure~\ref{fig:three_workloads} (left) visualizes the resource usage of each workload as a memory-latency box. When scheduled independently, these misaligned workloads leave significant \emph{holes} in GPU memory and timeline—leading to fragmentation and low utilization.
For instance, a retraining task may consume only a fraction of GPU memory but blocks the entire iteration window. Conversely, prioritizing inference may delay critical fine-tunes that personalize the model.
Inspired by continuous batching~\cite{yu2022orca}, we extend iteration-level scheduling to jointly colocate fine-tuning with prefill and decode. 
By dynamically adjusting batch sizes to opportunistically fill leftover memory within each iteration, the system mitigates fragmentation and exploits latency complementarity while respecting alignment freshness. Figure~\ref{fig:three_workloads} (right) illustrates this idea.

\begin{figure}
  \centering
  \begin{subfigure}[t]{\linewidth}
        \centering
        \includegraphics[width=\linewidth]{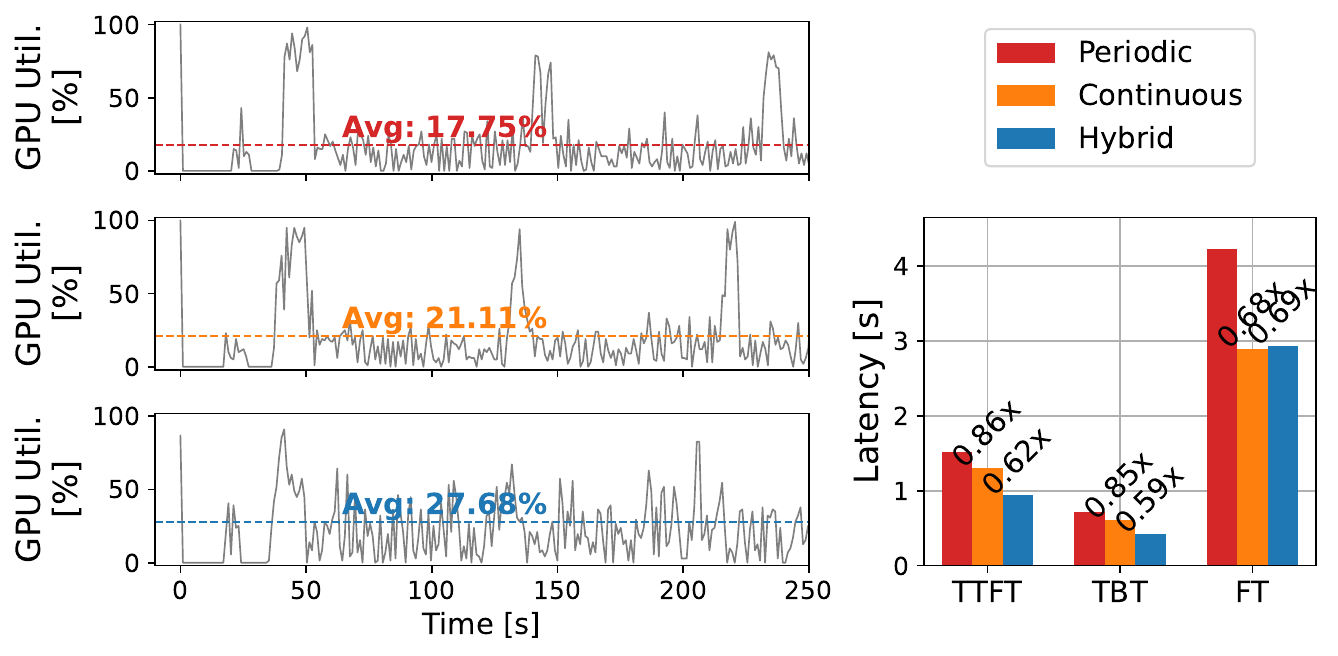}
        \caption{Mistral-7B on RTX A6000 Ada.} 
        \label{fig:util_server}
   \end{subfigure}
   \\
   \begin{subfigure}[t]{\linewidth}
        \centering
        \includegraphics[width=\linewidth]{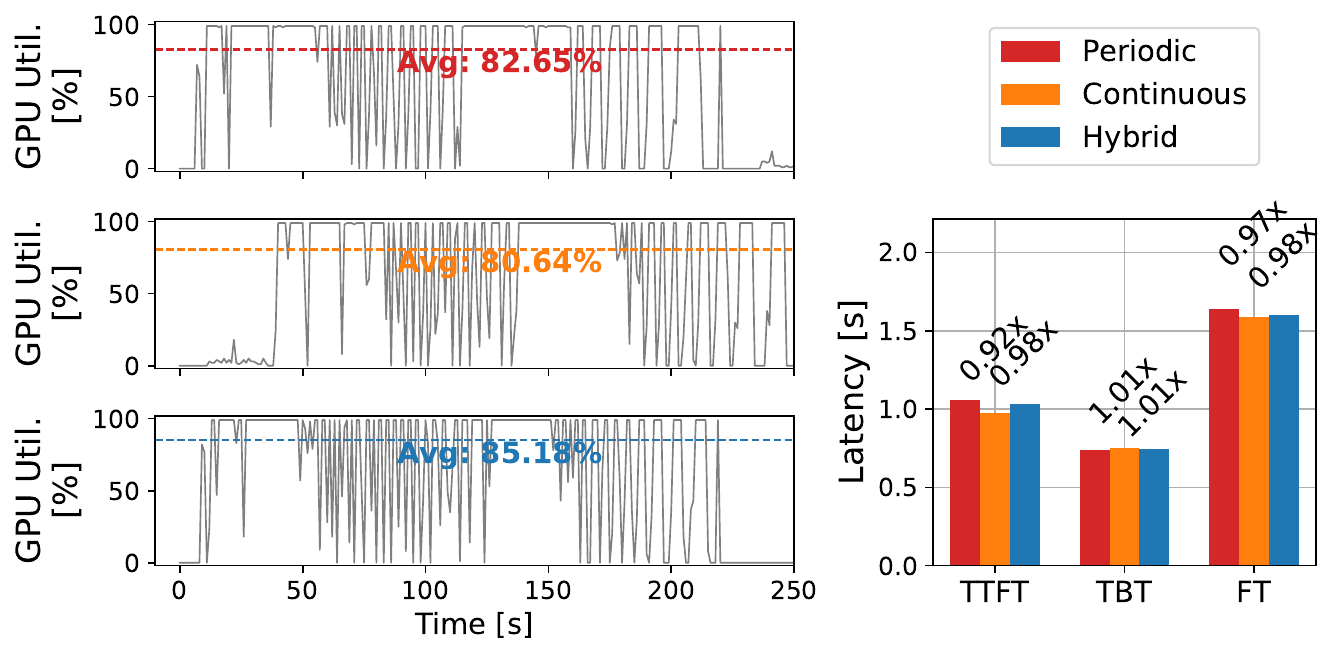}
        \caption{Mistral-3B on NVIDIA AGX Orin.} 
        \label{fig:util_orin}
    \end{subfigure}
    \caption{\emph{Left}: GPU utilization, and \emph{Right}: Latency breakdown on two (a) A6000 Ada server and (b) AGX Orin edge device.}
    \label{fig:util}
  \vspace{-5pt}
\end{figure}

\emph{\uline{A6000 server (Figure~\ref{fig:util_server})—fragmentation-dominated and hybrid gains}}.
On the A6000 server, average utilization is low: Periodic 17.8\% / Continuous 21.1\% / Hybrid 27.7\%. Here, the bottleneck is not compute saturation but fragmentation: alternating between training and inference leaves idle gaps in both time and memory. Continuous retraining partially fills these gaps, but Hybrid scheduling is far more effective by colocating small-batch fine-tuning with inference workloads and resizing batches per iteration. This directly translates into significant latency reduction: \emph{TTFT}: 0.62$\times$-0.86$\times$, \emph{TBT}: 0.59$\times$-0.85$\times$, and \emph{FT}: 0.68$\times$-0.86$\times$.

\emph{\uline{Orin (Figure~\ref{fig:util_orin})—compute-bound, consistently saturated}}.
On the AGX Orin edge device, the average utilization remains consistently high across all strategies: Periodic 82.7\% / Continuous 80.6\% / Hybrid 85.2\%. With tighter compute and memory budgets, any workload—particularly fine-tuning—pushes the GPU close to saturation. As a result, the opportunity to further increase parallelism is limited. The latency breakdown confirms this: relative to Periodic, Continuous and Hybrid only provide marginal improvements (TTFT 0.92$\times$-0.98$\times$, TBT$\sim$1.01$\times$, FT 0.97$\times$-0.98$\times$). In short, on compute-constrained edge platforms, scheduling refinements mainly ensure that continuous training does not worsen inference latency, while the overall improvement remains bounded by hardware ceilings.

\section{System Design of \approach}
\label{sec:method}

\begin{figure}[t]
    \centering
    \includegraphics[width=\linewidth]{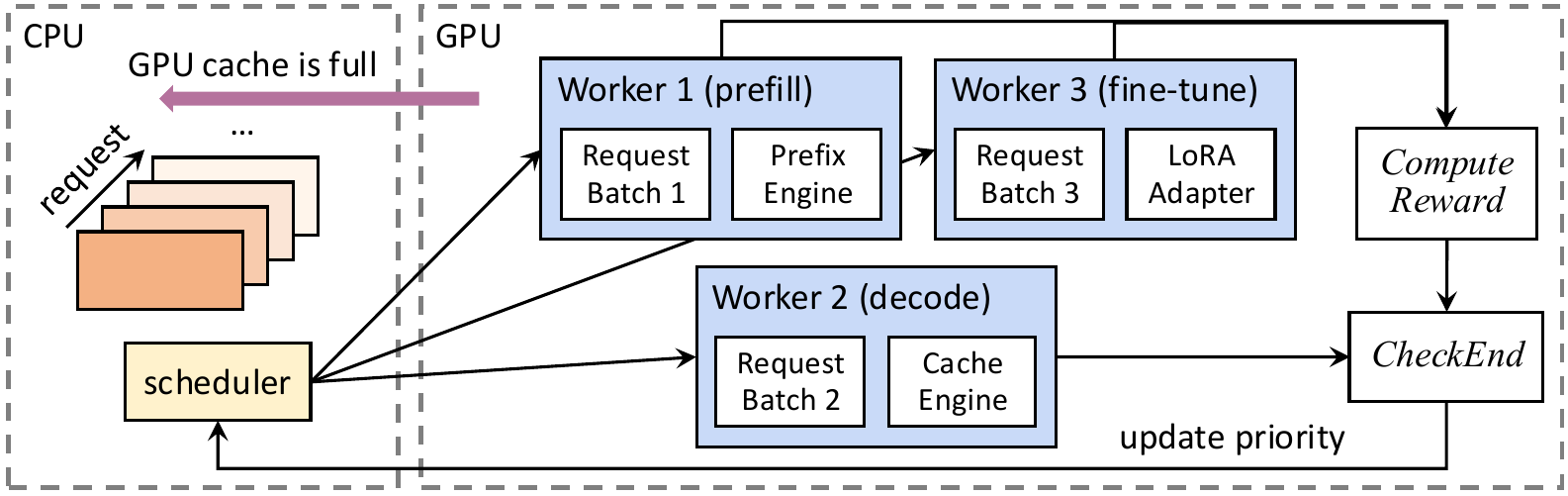}
    \caption{Design overview of \approach. The scheduler dispatches requests to prefill (prefix sharing), decode (KV cache pruning), and fine-tune (LoRA) workers, while feedback from reward computation and queuing time enables real-time priority updates under CPU–GPU cache coordination.}
    \label{fig:overview}
    \vspace{-10pt}
\end{figure}

We present \approach, a fine-grained GPU-local scheduling system designed to support colocated LLM inference serving and continuous retraining. Figure~\ref{fig:overview} illustrates the design: user requests arrive asynchronously into a shared priority queue. The scheduler dynamically assigns priorities and batches them into iterations using a best-fit bin packing strategy, accounting for memory fragmentation and alignment reward. These batches are dispatched to GPU workers, where each worker concurrently handles prefilling, decoding, or fine-tuning workloads. After execution, tasks are re-prioritized if not finished and pushed back into the queue.
\approach is built on three key components:
\begin{itemize}[nosep,leftmargin=*]
    \item \textbf{Priority computation:} Assign dynamic priority to each request based on workload type, enqueued time, and accuracy-driven rewards (e.g., DPO loss).
    \item \textbf{Iteration-level memory-aware batching:} A GPU-local scheduler that allocates tasks to minimize memory fragmentation and latency interference.
    \item \textbf{Cache management:} Optimizes memory reuse through prefix sharing in prefilling and cache pruning in decoding, and evicts unused cache to prevent memory bloat.
\end{itemize}

\subsection{Alignment-aware Prioritization}
\label{sec:priority}

Each incoming request is tagged with its arrival time $t_\text{arr}$ and workload type $\mathsf{w} \in \{\mathsf{prefill}, \mathsf{decode}, \mathsf{ft}\}$. We define its base priority $\pi_\mathsf{w}$ and temporal growth rate $\delta_\mathsf{w}$, allowing a dynamic priority $P(t)$ at current time $t$ to grow over time:
\[
P_\mathsf{w}(t) = \pi_\mathsf{w} + \delta_\mathsf{w} \cdot (t - t_\text{arr}),
\]
where we set $\pi_\mathsf{decode} > \pi_\mathsf{prefill} > \pi_\mathsf{ft}$ to prioritize inference serving and avoid SLO violation caused by frequent preemption, and set $\delta_\mathsf{ft} > \delta_\mathsf{prefill} > \delta_\mathsf{decode}$ to gradually escalate the criticality of fine-tuning requests over time to avoid \emph{retraining starvation}. 

Additionally, to promote retraining samples that are particularly valuable for model alignment, \approach integrates their alignment reward (i.e., DPO loss) to the priority:
\[
P_{\mathsf{ft}}^{\text{total}}(t) = P_{\mathsf{ft}}(t) + \gamma \cdot \mathcal{L}_{\text{DPO}}(x, y^+, y^-),
\]
where $\gamma$ is a tunable weight. This mechanism prioritizes retraining samples with higher misalignment. As shown in Figure~\ref{fig:bin_packing}, after prioritization at iteration 9, the tasks are ordered by priority as $A_{ft}$, $C_{d}$, $F_{d}$, $E_{d}$, $H_{p}$, $G_{d}$, and $D_{ft}$ (from highest to lowest).

\subsection{Iteration-level Resource-aware Batching}
\label{sec:bin_packing}

\begin{algorithm}[t]
\caption{GPU-Local Request Scheduling}
\label{alg:bestfit}
\small
\begin{algorithmic}[1]
\REQUIRE Task queue $\mathcal{Q}$, memory capacity $\mathcal{P}$, thresholds $(\tau_\text{mem}, \tau_\text{task})$, memory-latency weights $(\lambda_1, \lambda_2)$.
\ENSURE Scheduled bin $\mathcal{B}_1$ for execution.

\STATE Initialize bin list $\mathcal{B} \gets []$, task counter $c \gets 0$
\WHILE{$\mathcal{Q}$ not empty}
    \IF{$\mathcal{B} \neq \emptyset \wedge (\mathcal{B}_1.\text{memory} \geq \tau_\text{mem}\cdot \mathcal{P} \vee c \geq \tau_\text{task})$}
        \STATE \textbf{break}
    \ENDIF
    \STATE Retrieve $\text{task} \gets \mathcal{Q}$.dequeue(), increment $c \gets c+1$
    \STATE Estimate workload: $(m, \ell) \gets \textsc{GetWorkload}(\text{task})$
    \STATE Initialize $\text{best\_bin} \gets \texttt{Null}$, $\text{best\_score} \gets \infty$
    \FOR {each bin $\mathcal{B}_i$ in $\mathcal{B}$}
        \IF{$\mathcal{B}_i.\text{free\_memory} \geq m$}
            \STATE Compute fragmentation score $f$
            \IF{$\text{score} < \text{best\_score}$}
                \STATE Update: $\text{best\_score} \gets \text{score}, \text{best\_bin} \gets \mathcal{B}_i$
            \ENDIF
        \ENDIF
    \ENDFOR
    \IF{$\text{best\_bin} \neq \texttt{None}$}
        \STATE Assign task to $\text{best\_bin}$, update memory and latency
    \ELSE
        \STATE Create new bin $\mathcal{B}_{\text{new}}$, insert task, append to $\mathcal{B}$
    \ENDIF
\ENDWHILE

\IF{$\mathcal{B} \neq \emptyset$}
    \FOR{$i = 2$ to $|\mathcal{B}|$}
        \FOR{each task in $\mathcal{B}_i$}
            \IF{$\textsc{CheckEnd}(\text{task}) == \texttt{False}$}
                \STATE Push task back into $\mathcal{Q}$ with updated priority
            \ENDIF
        \ENDFOR
    \ENDFOR
\ENDIF

\STATE \textbf{Output:} $\mathcal{B}_1$
\end{algorithmic}
\end{algorithm}

\noindent\textbf{Best-fit heuristic motivation.} Scheduling a mix of LLM workloads on a single GPU can be viewed as a bin-packing problem in which each iteration’s “bin” is the available GPU memory and each request (prefill, decode, ft) is an execution unit with particular sizes (memory footprint) and duration (latency). Unlike general static bin packing, \approach involves an online, heterogeneous stream of tasks with different memory/latency profiles (Figure~\ref{fig:latency_memory_profile}). A naive greedy scheduler that runs tasks sequentially or without regard for workload size can lead to low GPU utilization—e.g. large training jobs leaving gaps of unused memory that smaller inference jobs could have filled, or short decoding tasks waiting behind long-running jobs. 

\approach adopts a best-fit heuristic as a pragmatic solution to this dynamic packing problem. Best-fit scheduling greedily fills the GPU with the combination of tasks that most tightly fits the available memory, while also accounting for execution time compatibility. It minimizes memory fragmentation (unutilized memory, which appears as the red gap in Figure~\ref{fig:three_workloads}) and helps balance latency so that no single task unduly delays the others. Formally, the scheduler assigns a composite score $f$ to each candidate set of tasks using a two-dimensional objective:
\[
f = \lambda_1 \cdot | \mathcal{B}_i.\text{free\_memory} - M | + \lambda_2 \cdot | \mathcal{B}_i.\text{max\_latency} - \ell |,
\]
where $\lambda_1, \lambda_2$ trade off vertical and horizontal memory fragmentation. In practice, this best-fit heuristic is fast, adaptive, and well-suited to colocated workloads, even though globally optimal packing is NP-hard. By prioritizing a tight memory fit, we ensure that the GPU’s capacity is used as much as possible each iteration, while the latency-aware term prevents pathological cases (e.g. packing a very slow task with many fast tasks that would all finish and leave the GPU underutilized waiting for the slow one).

\begin{figure}
    \includegraphics[width=\linewidth]{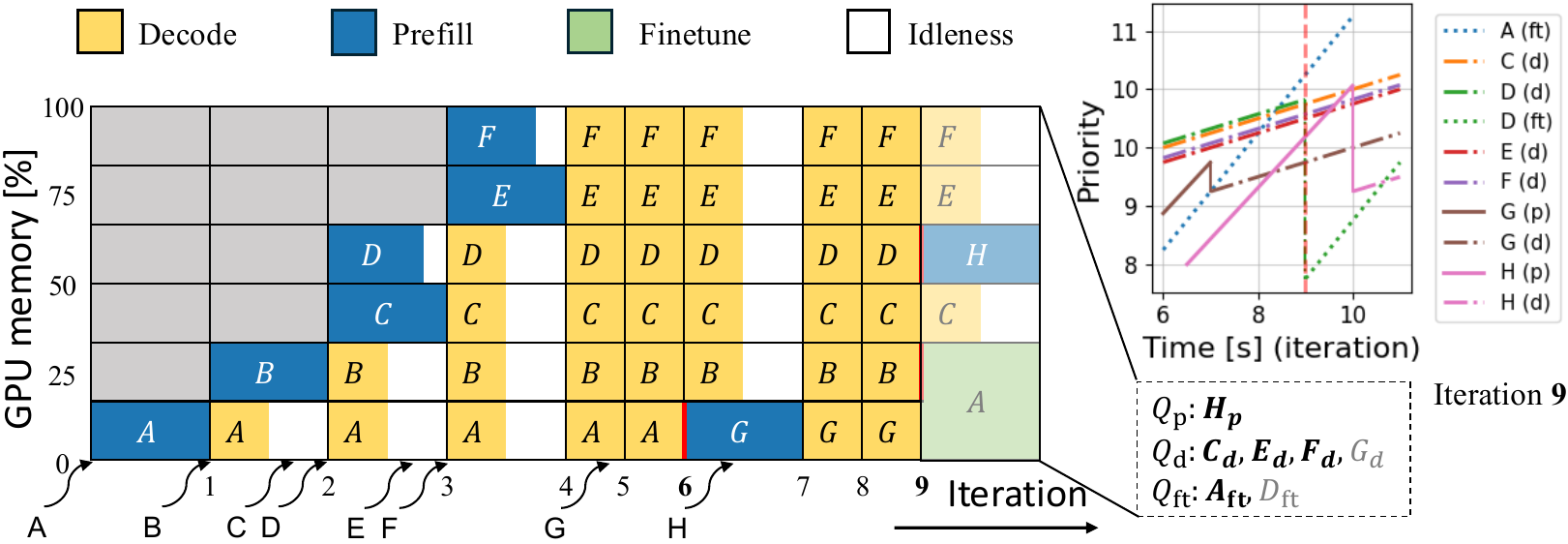}
    \caption{GPU-local scheduling of inference and training requests. $Q_p$, $O_d$, $Q_{\text{ft}}$ denote queues for prefill, decode, and fine-tune. \textbf{Bold} texts denote scheduled requests for the next bin (iteration). }
    \label{fig:bin_packing}
    \vspace{-10pt}
\end{figure}

\noindent\textbf{Bin allocation.}
As illustrated in Algorithm~\ref{alg:bestfit}, the scheduler dequeues tasks sorted by dynamic priority (\S\ref{sec:priority}) and attempts to assign them to the current iteration bin using the best-fit score (lines 1-16). It considers memory fit to reduce fragmentation and latency divergence to avoid stalls. This enables effective co-scheduling of short decode jobs and large fine-tune jobs within a single GPU iteration. After execution, the system decides whether to reschedule unfinished tasks (lines 17–21) according to the \texttt{CheckEnd} output:
\begin{itemize}[nosep, leftmargin=*]
    \item For inference, it determines if the next decoded token is [EOS] or if the decoding iteration reaches the predefined maximum steps. 
    \item For retraining, it checks whether DPO loss is still above a pre-defined threshold. 
\end{itemize}
Tasks needing further iterations are requeued with updated priorities; others are retired.
Figure~\ref{fig:bin_packing} demonstrates the effectiveness of \approach. In iteration~6, a bin is formed by packing several decode jobs $B_d\sim F_d$ and one prefill $G_p$, with a large fine-tune $A_{ft}$ temporarily delayed to prioritize throughput. While in iteration~9, the scheduler flexibly prioritizes the fine-tune $A_{ft}$ which now has a higher priority, and delays the decode $G_d$ to ensure timely model updates. This memory-aware batching balances the idleness-dependent throughput and continuous alignment learning process for heterogeneous workloads.

\subsection{Prefix Sharing and Cache Management}
\label{sec:cache_management}

To further mitigate the retraining–inference tradeoff, \approach leverages two key optimization techniques: \emph{Prefix sharing}~\cite{lin2024parrot,zheng2024sglang} and \emph{KV cache pruning}~\cite{li2024snapkv,fu2025not}. These techniques aim to reduce memory and latency overhead for inference, thereby creating more opportunities for retraining without compromising serving throughput.

\begin{figure}[]
    \centering
    \begin{subfigure}[t]{0.485\linewidth}
        \centering
        \includegraphics[width=\linewidth]{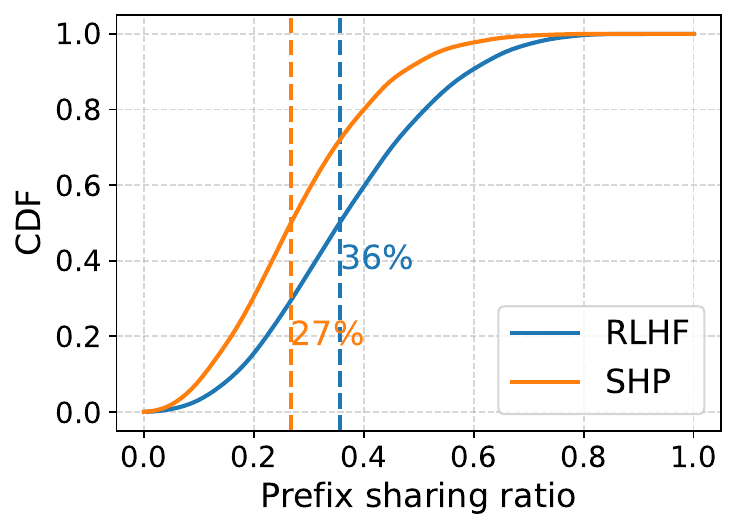}
        \caption{Prefix sharing in prefill}
        \label{fig:prefill_ttft}
    \end{subfigure}
    \hfill
    \begin{subfigure}[t]{0.49\linewidth}
        \centering
        \includegraphics[width=\linewidth]{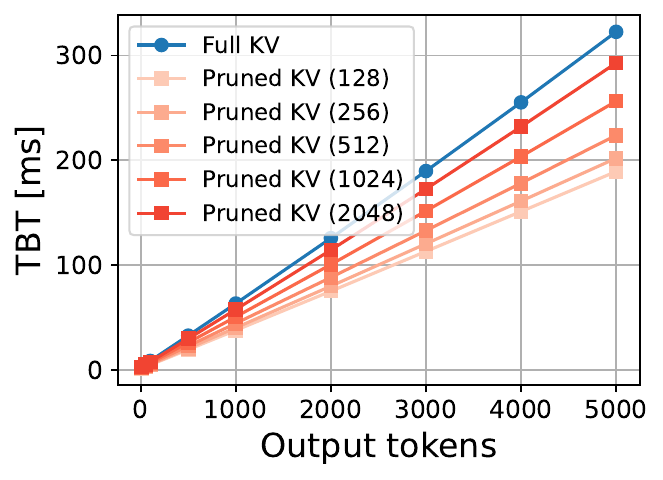}
        \caption{Cache pruning in decode}
        \label{fig:decoding_tbt}
    \end{subfigure}
    \caption{Prefix sharing ratio during prefill and decoding TBT of pruned KV cache with varying capacity size $C_{\text{total}}$.}
    \vspace{-10pt}
    \label{fig:cache_latency}
\end{figure}

\noindent\textbf{Reduced prefill via prefix tree.}
Prefix sharing exploits input redundancy across user queries to avoid redundant prefill computation. As illustrated in Appendix~\ref{app:method}, multiple user queries often share common token prefixes, e.g., ``\emph{What is the best way to...}''. We construct a prefix tree over active inference requests, where each leaf represents a complete request and each internal node represents a shared prefix segment. This design follows a Trie structure~\cite{zheng2024sglang}, where a path from the root to any leaf captures the longest common prefix of a request. To maximize reuse, we traverse the tree in a DFS manner to co-execute shared prefixes.
At each internal node of the prefix tree, we cache the KV pairs from the prefill outputs. This cache is efficiently reused by descendant nodes without recomputation. To manage memory usage, we offload least recently used (LRU) cache when inference concurrency is high to release more memory budget. By avoiding repeated computation for shared prefix tokens, prefix sharing reduces both latency and memory consumption in prefill. It also provides more headroom for opportunistic retraining, as the scheduler can pack training tasks without disrupting inference throughput.
Figure~\ref{fig:prefill_ttft} demonstrates the distribution of prefix sharing ratios in two benchmark datasets.

\noindent\textbf{Decode KV cache pruning.}
Decoding presents another challenge: KV cache for each output token accumulates linearly and quickly saturates GPU memory. To mitigate this, we introduce a norm-based cache pruning mechanism, inspired by the revealed contextual sparsity of LLMs~\cite{liu2023deja}.
At each decoding iteration, we compute the L2-norm of the output attention vector across heads. Let $\mathbf{a}_{t, h} \in \mathbb{R}^d$ denote the output at iteration $t$ and attention head $h$, and define its norm as $\left\|\mathbf{a}_{t, h}\right\|_2$. We maintain a rolling average $\bar{\mathbf{a}}_h$ for each head and reallocate per-head KV storage capacity proportionally:
\[
\bar{\mathbf{a}}_h = \frac{1}{T} \sum_{t=1}^{T} \left\| \mathbf{a}_{t, h} \right\|_2,~w_h = \frac{\bar{\mathbf{a}}_h}{\sum_j \bar{\mathbf{a}}_j},~C_h = \left\lfloor w_h \cdot C_{\text{total}} \right\rfloor.
\]
Here, $\bar{\mathbf{a}}_h$ is the average $\ell_2$ norm of the attention output at head $h$, $C_{\text{total}}$ is the total available KV cache capacity, and $C_h$ is the per-head allocation. The remaining capacity is greedily assigned to heads with the largest weight $w_h$:
\[
\text{Prune if} \quad t - t_{\text{last\_used}} > W \quad \vee \quad \left\| \mathbf{a}_{t, h} \right\|_2 < \tau,
\]
where $W$ is a fixed sliding window size and $\tau$ is a norm threshold below which attention heads are considered uninformative~\cite{fu2025not}.
This dynamic reallocation selectively retains high-signal attention heads while pruning less informative ones. If a head’s norm is consistently small, it receives lower cache budget and may prune earlier attention positions. As shown in Figure~\ref{fig:decoding_tbt}, this reduces decoding latency and is helpful under long outputs or memory pressure.
By combining prefix sharing for prefill and norm-guided cache pruning for decoding, we allow more training jobs to execute without sacrificing inference throughput. 

\section{Discussion and Implementation}
We build our system by integrating efficient LLM inference engine, lightweight retraining modules, and a scheduler with fine-grained concurrency and batching support. Below, we elaborate on each component.

\noindent\textbf{Inference engine.} 
Our system adopts vLLM~\cite{kwon2023efficient}, an inference backend optimized for decoding throughput using PagedAttention and FlashAttention~\cite{dao2022flashattention} mechanisms. This enables our system to leverage token-level parallelism and key-value memory optimization for high-throughput generation.


\noindent\textbf{Personalized adapter.}
To enable low-latency retraining without degrading inference throughput, we leverage PEFT techniques by assigning each user or domain an independent LoRA adapter $\phi_u$~\cite{hu2022lora}. This allows user-personalized updates while sharing the frozen base model $\theta$ across all requests, enabling tighter integration with online serving in each shared GPU environment. 

\noindent\textbf{Concurrency support.} 
\approach is implemented with multi-threaded concurrency using a thread pool design: A \emph{producer} thread ingests inference/retraining requests. A \emph{scheduler} thread computes dynamic priorities and forms hybrid iteration-level batches, where a bin allocator maps scheduled jobs to physical GPU sessions. Multiple \emph{executor} threads run on-device workloads asynchronously.
To balance latency and throughput, the scheduler maintains a priority queue that supports: \emph{Dynamic priority refresh}: periodically updates request priorities based on utility functions (e.g., model freshness, deadline slack). \emph{Preemption-aware iteration batching}: schedules batches with least .

\section{Evaluation}
\label{sec:evaluation}

\subsection{Experimental Setup}

\noindent\textbf{Hardware platforms.} 
We test both \emph{server} with NVIDIA A6000 Ada GPU (48 GB VRAM) and \emph{edge} device with NVIDIA AGX Orin (32GB  LPDDR5), ensuring a controlled environment to isolate the scheduler's effects. 

\noindent\textbf{Workloads and datasets.} 
We simulate a mixed workload of inference and retraining requests representing single- and multiple-user domains with varying alignment needs. Inference requests are generated as a Poisson arrival process~\cite{li2023rt}, and additional fraction of them (i.e., retrain rate, which is typically less than 50\%~\cite{bhardwaj2022ekya}) are retraining jobs. We consider two datasets to emulate different alignment domains or user preferences:
\begin{itemize}[nosep, leftmargin=*]
    \item \textbf{RLHF}~\cite{bai2022training} (\emph{harmlessness}): Human feedback focusing on safety alignment—preferred responses tend to avoid unsafe or illegal content. This represents users requiring the model to refuse harmful outputs. 
    \item \textbf{SHP}~\cite{ethayarajh22a} (\emph{helpfulness}): A broad human preference dataset with 385k comparisons across 18 topics, emphasizing helpfulness of responses over others. This represents general user satisfaction alignment.
\end{itemize}
By using these distinct datasets, we create multi-tenant scenarios where different ``tenants'' (user groups) issue requests with different alignment objectives. This diversity lets us evaluate \approach’s overall alignment capability.

\noindent\textbf{Baselines.}
We compare \approach against several baseline scheduling strategies:
\begin{itemize}[nosep, leftmargin=*]
    \item \textbf{Ekya}~\cite{bhardwaj2022ekya}: A policy that periodically preempts inference to run retraining at fixed intervals. Inference requests queue up during retraining epochs (synchronous blocking). This baseline represents traditional FT where serving and training are separated in phases.
    \item \textbf{AdaInf}~\cite{shubha2023adainf}: Continuous scheduling that gives retraining jobs highest priority in a FIFO queue, where incoming FT tasks can preempt or delay inference until they complete. It maximizes retraining frequency but with no special handling for interference, e.g., often hurting inference latency significantly.
\end{itemize}




\subsection{Performance under Varying Retraining Intensity}

We evaluate \approach against Ekya and AdaInf under varying retraining intensities, focusing on alignment accuracy, inference throughput, and latency breakdown.

\begin{figure}
    \centering
    \includegraphics[width=0.9\linewidth]{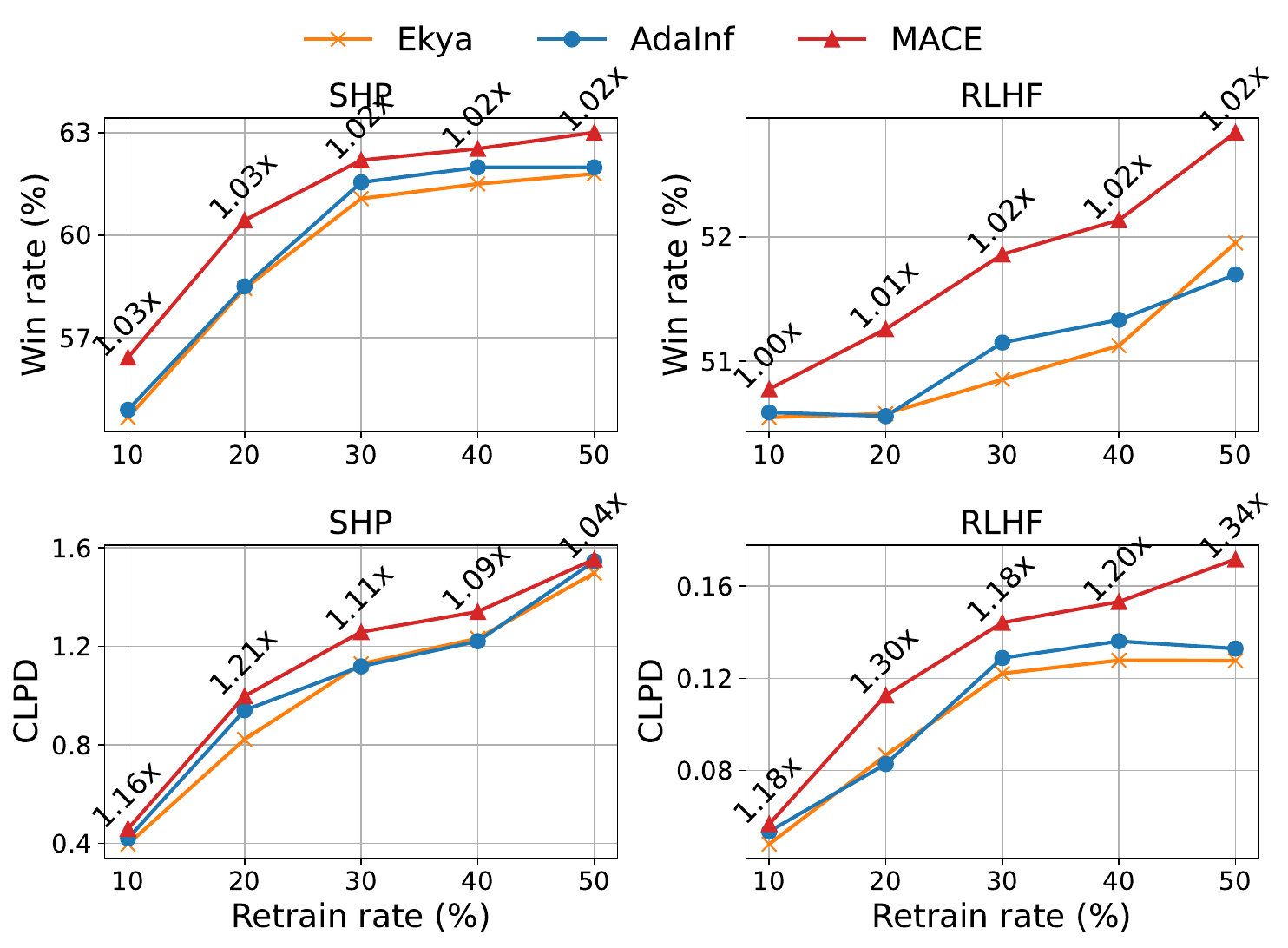}
    \caption{Average win rate and CLPD across various retrain rates of different methods on \emph{Left}: SHP and \emph{Right}: RLHF dataset.}
    \label{fig:main_accuracy}
    \vspace{-10pt}
\end{figure}

\noindent\textbf{Alignment accuracy.}
Figure~\ref{fig:main_accuracy} shows average win rate and CLPD across SHP and RLHF datasets. \approach consistently outperforms both baselines, especially under higher retraining rates. On SHP, \approach achieves up to 1.03$\times$ higher win rate and 1.21$\times$ higher CLPD compared to AdaInf. On RLHF, the advantage is more pronounced: \approach improves win rate by up to 1.02$\times$ and CLPD by 1.34$\times$, reflecting its ability to retain alignment accuracy even under intensive retraining. In contrast, Ekya and AdaInf show diminishing returns as retraining frequency increases, highlighting the importance of \approach’s fine-grained scheduling and cache-aware optimizations.

\begin{figure}
  \centering
  \begin{subfigure}[t]{\linewidth}
        \centering
        \includegraphics[width=0.9\linewidth]{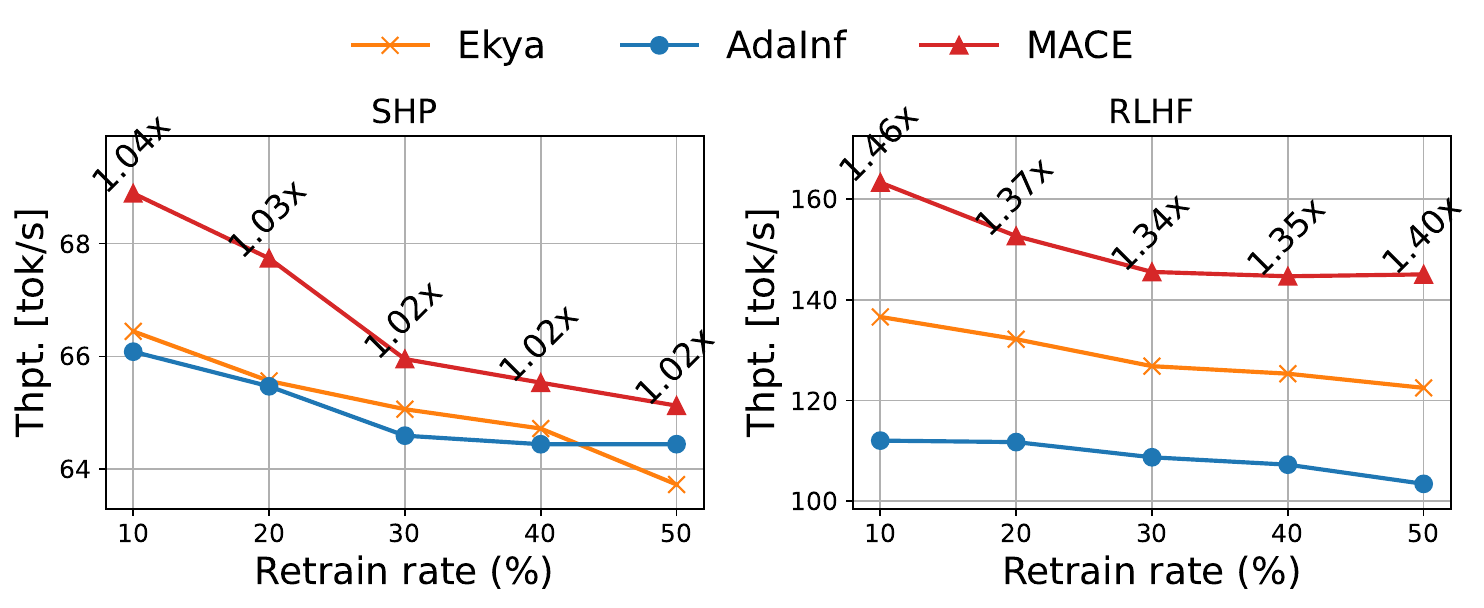}
        \caption{Throughput (token/sec) of Mistral-7B on RTX A6000 Ada.} 
        \label{fig:throughput_server}
   \end{subfigure}
   \\
   \begin{subfigure}[t]{\linewidth}
        \centering
        \includegraphics[width=0.9\linewidth]{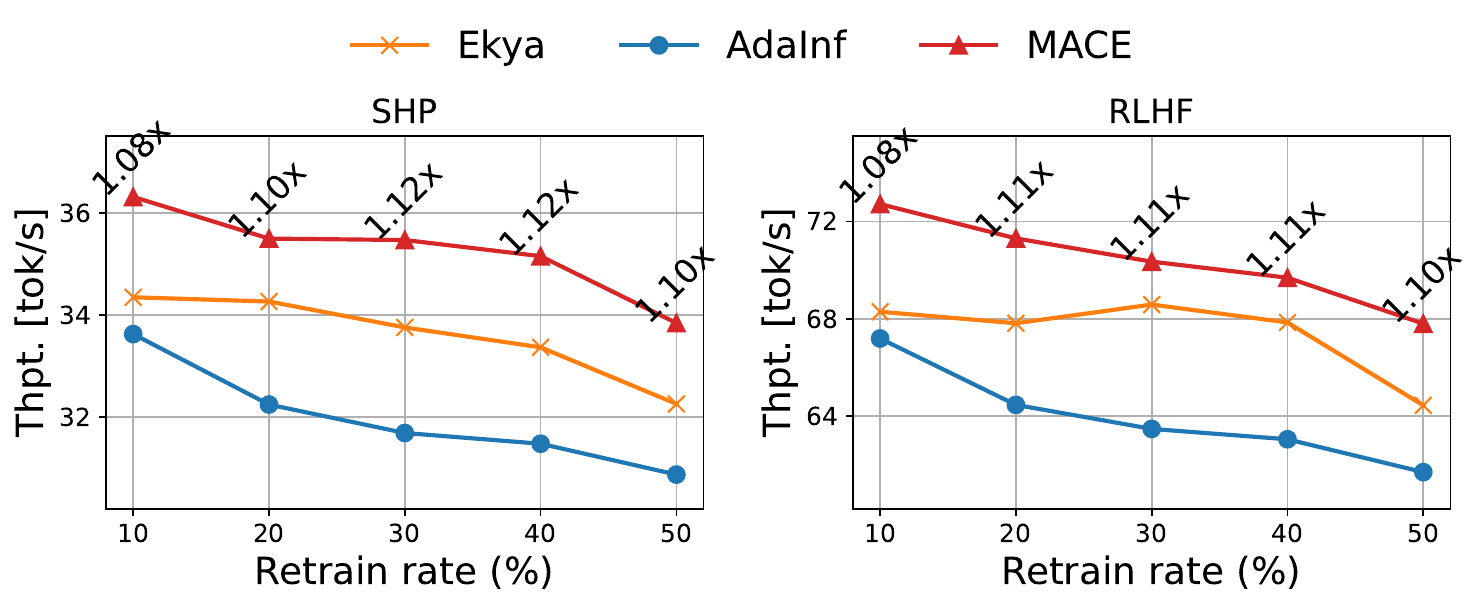}
        \caption{Throughput (token/sec) of Mistral-3B on NVIDIA AGX Orin.} 
        \label{fig:throughput_orin}
    \end{subfigure}
    \caption{Inference throughput comparison of different methods.}
    \label{fig:main_throughput}
  \vspace{-5pt}
\end{figure}

\noindent\textbf{Throughput.}
Figure~\ref{fig:main_throughput} reports inference throughput on both server-scale and edge-scale platforms. On the A6000, \approach delivers up to 1.46$\times$ higher throughput than AdaInf at 10\% retraining rate, while maintaining superior accuracy. On the AGX Orin, which is more resource constrained, \approach sustains 1.12$\times$ higher throughput on average, with benefits most evident at higher retraining intensities where baselines degrade rapidly. These results demonstrate that \approach’s hybrid batching and pruning strategies effectively amortize retraining costs without sacrificing inference performance, a critical property for deployment on heterogeneous hardware.

\begin{figure} 
\centering 
    \begin{subfigure}[t]{0.49\linewidth} 
    \includegraphics[width=\linewidth]{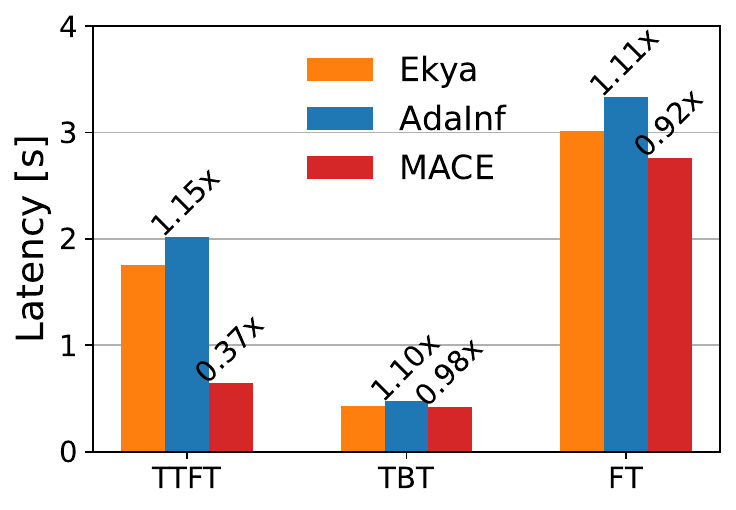}       
    \caption{\footnotesize RTX A6000 Ada}           \label{fig:latency_breakdown_server} \end{subfigure} 
\hfill 
    \begin{subfigure}[t]{0.49\linewidth} 
    \centering 
    \includegraphics[width=\linewidth]{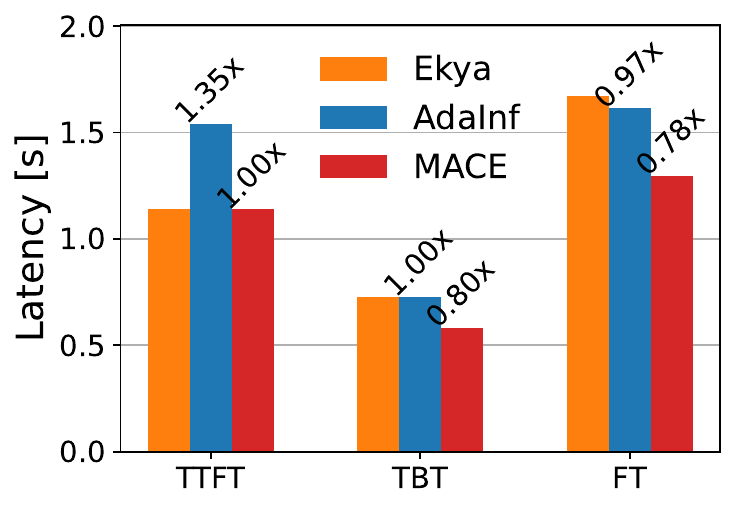} \caption{\footnotesize NVIDIA AGX Orin} \label{fig:latency_breakdown_orin} \end{subfigure} 
    \caption{Latency breakdown on (a) A6000 and (b) AGX Orin.} 
\label{fig:main_latency_breakdown} 
\end{figure}

\noindent\textbf{Latency breakdown.}
Figure~\ref{fig:main_latency_breakdown} decomposes end-to-end latency into prefill, decode, and fine-tuning components. \approach achieves lower decode latency (TBT) via cache pruning, while prefix sharing substantially reduces prefill cost (TTFT). Fine-tuning overhead remains bounded due to iteration-level batching, which overlaps retraining with inference execution. Compared to AdaInf, \approach reduces prefill latency by 63\% on A6000 server and decode latency by 30\% on edge device platforms. This balanced reduction explains its ability to simultaneously improve alignment accuracy and throughput, highlighting the effectiveness of joint scheduling across retraining and inference phases.


\subsection{Understanding \approach}


To better understand the contribution of each component in \approach, we create the following variants by disabling specific mechanisms:
\begin{itemize}[nosep, leftmargin=*]
    \item \emph{MACE/Bin}: This variant disables memory-aware batching by fixing maximum batch sizes and applying FIFO queuing without hybrid scheduling.
    \item \emph{MACE/Prefix}: This variant disables prefix sharing during the prefill phase, resulting in redundant computation across similar requests.
    \item \emph{MACE/Prune}: This variant disables KV cache pruning under memory constraints, forcing full cache retention during decoding.
\end{itemize}


\begin{figure}
    \centering
    \includegraphics[width=\linewidth]{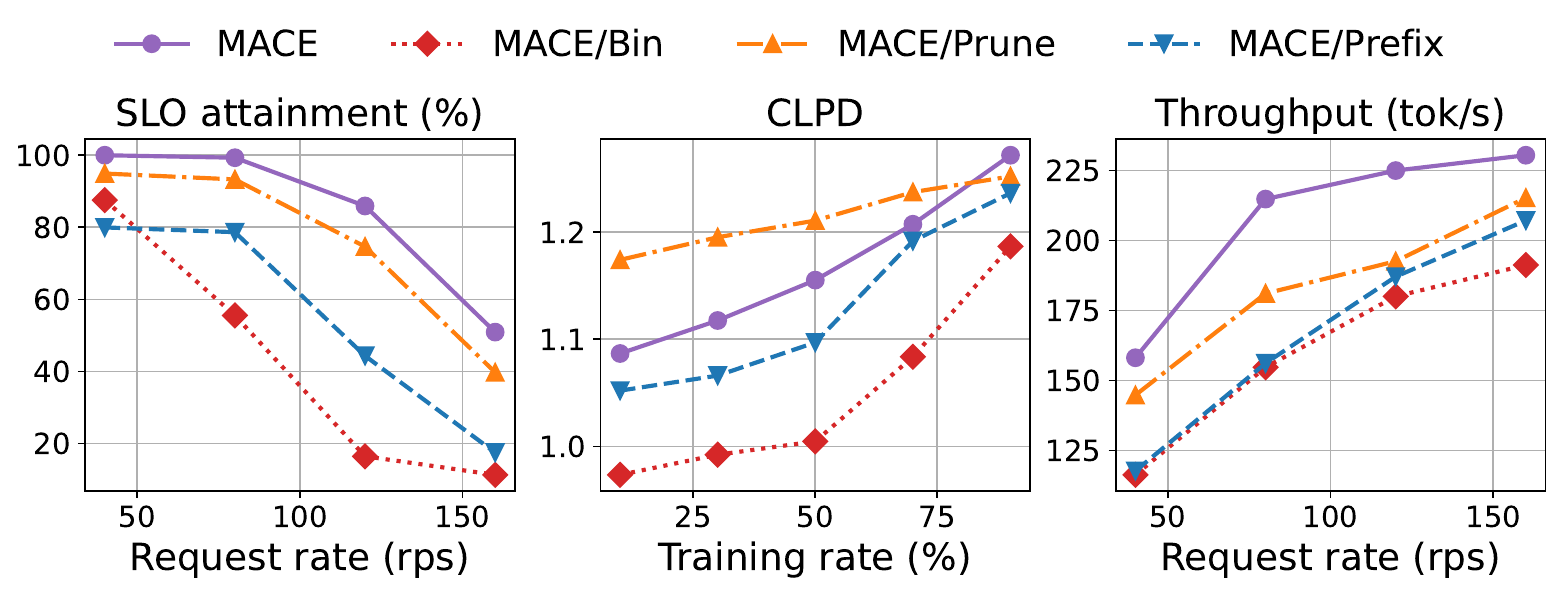}
    \caption{\emph{Left}: SLO attainment rate, \emph{Middle}: CLPD in SHP, and \emph{Right}: inference throughput of different variants of \approach.}
    \label{fig:ablation}
\end{figure}

\noindent\textbf{Ablation study.}
We measure SLO attainment—proportion of inference tasks whose TTFT meet a SLO deadline of 5× forward latency~\cite{li2023alpaserve}. Figure~\ref{fig:ablation} (left) shows the SLO attainment rate under increasing request loads.
Disabling hybrid scheduling (MACE/Bin) causes the steepest decline, with attainment falling below 20\% at high load, confirming that fixed batching fails to adapt to workload burstiness. MACE/Prefix also shows a 35–40\% lower attainment than full \approach at 150 rps, as redundant prefill computations delay request completion. These results highlight that hybrid scheduling and prefix sharing are key to meeting latency SLOs under heavy traffic.
On alignment accuracy (CLPD) in Figure~\ref{fig:ablation} (middle), \approach achieves up to 8\% improvement over MACE/Prefix and 15\% over MACE/Bin, showing the importance of dynamic batching and prefix sharing in reducing latency-induced drift. Particularly, MACE/Prune consistently shows the highest CLPD scores, indicating that pruning KV cache will sacrifice accuracy to a certain point.
For throughput, \approach delivers over 225 tok/sec, compared to drops of 7–11\% in MACE/Prune and MACE/Prefix, and 17.6\% in MACE/Bin. These results confirm that all three mechanisms contribute complementary benefits: hybrid scheduling is most critical for latency and accuracy, while prefix sharing and cache pruning jointly improve throughput.

\begin{figure}
    \centering 
    \includegraphics[width=\linewidth]{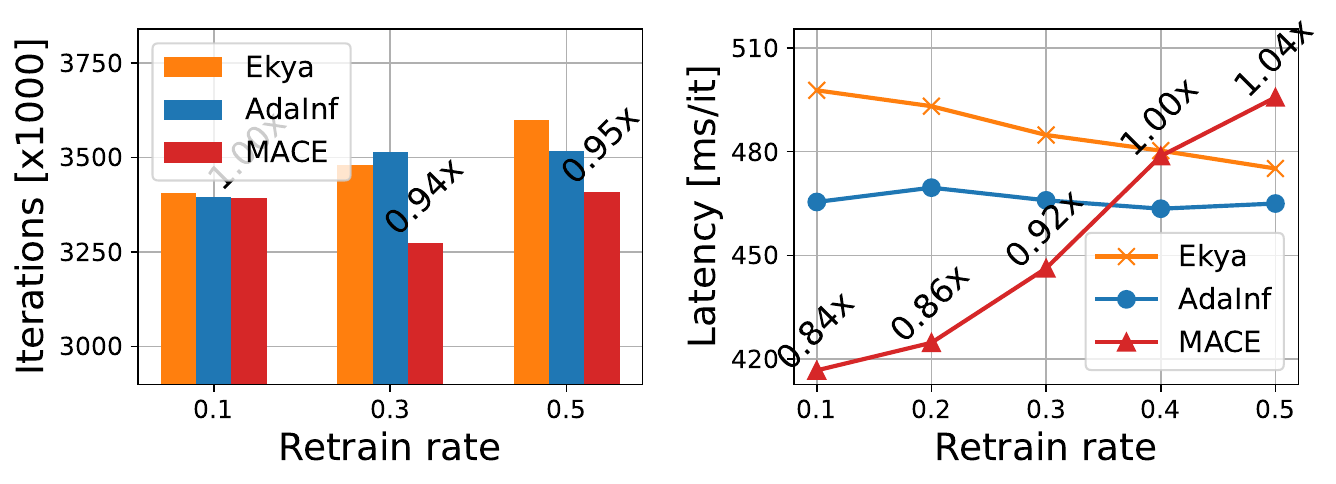}
    \caption{\emph{Left}: total iterations and \emph{Right}: average per-iteration latency for Mistral-7B on A6000 Ada.}
    \label{fig:latency_iteration}
    \vspace{-5pt}
\end{figure}

\noindent\textbf{Bin efficiency.}
Figure~\ref{fig:latency_iteration} shows that under the same concurrency constraints (e.g., inference batch size $\leq$ 50, train $\leq$ 2) and queuing policy (e.g., FIFO), hybrid requires fewer iterations to finish the same workload, especially when retraining is light. This suggests a better iteration efficiency by reducing idle memory slots.
Despite mixing heterogeneous workloads, its per-iteration latency remains close to baselines, indicating that hybrid effectively fills memory and latency gaps (i.e., fragmentation) without incurring extra runtime cost.

\begin{table}[]
    \centering
    \caption{Average time overhead (ms) and memory (MB) of schedulers. \sout{PA}, \sout{IB}, \sout{CM} mean without prioritization~(\S\ref{sec:priority}), iteration-level batching~(\S\ref{sec:bin_packing}), and cache management~(\S\ref{sec:cache_management}).
    } 
    \label{tab:profiling}
    \vskip 0.15in
    \resizebox{0.47\textwidth}{!}{
    \begin{tabular}{lccccc}
    \hline
       & \multirow{2}{*}{Ekya} & \multirow{2}{*}{AdaInf} & \multicolumn{3}{c}{{\approach}} \\
    \cline{4-6}
        & & & \sout{PA} & \sout{IB} & \sout{CM} \\
    \hline
      Latency (ms) & 3e-2 & 8e-2 & 1.1e-1  & 1.2e-1 & 1.5e-1 \\
    \hline
      Memory (MB) & 0.6 & 2 & 3.5 &  4.5 & 4  \\
    \hline
    \end{tabular}
    }
    \vspace{-10pt}
\end{table}

\noindent\textbf{Overhead analysis.}
As shown in Table~\ref{tab:profiling}, the added overhead is computationally lightweight and memory efficient (under 5MB total), and the latency cost is amortized across each batch. In real workloads with high GPU compute cost, this overhead is a worthwhile tradeoff for improved throughput and accuracy.
The scheduling decision in our full method introduces $\sim$15 ms/request overhead, about 5$\times$ higher than Ekya, which is expected given our scheduler tracks per-request priority, memory fragmentation estimates, and iteration-level packing. However, this is still negligible ($<$0.1\%) compared to typical request execution time (which ranges from tens to hundreds of milliseconds).
Compared to AdaInf, \approach adds $\sim$7 ms/request additional overhead due to more fine-grained request-level decision-making instead of coarse DAG session-level scheduling.
Our full scheduler uses up to 5MB, largely for per-iteration job queues, lightweight memory pressure estimation, and cache tracking structures (prefix/KV/retention flags). In contrast, Ekya and AdaInf maintain only coarse batch/session metadata (no fine-grained tracking), hence use $<$2MB.





\section{Limitations and Discussion}
\noindent\textbf{Scalability beyond a single node.}
Our current design and evaluation focus on a single edge server equipped with limited GPU resources. While the method naturally generalizes to multiple concurrent models and tasks within a node, scaling across multiple edge nodes or hybrid cloud-edge architectures introduces challenges such as synchronization of model versions, cross-node retraining coordination, and distributed cache consistency. Future extensions may explore hierarchical or federated variants of \approach that operate over distributed GPU clusters.

\noindent\textbf{GPU heterogeneity and offline profiling.}
Our current system assumes access to a single GPU type within a node, following prior work~\cite{romero2021infaas,shubha2023adainf}. Extending \approach to support heterogeneous GPU types or multi-node edge clusters introduces new challenges in profiling transfer latency, maintaining cache coherency, and synchronizing retraining checkpoints across devices. Additionally, as with prior systems, we rely on offline profiling to characterize prefill/decode latencies under different batch sizes and context lengths. Reducing this profiling overhead or making the scheduler online-adaptive remains an important direction for future work.

\noindent\textbf{CPU execution alternatives.}
Several inference systems consider fallbacks to CPU execution under low-load scenarios~\cite{guo2021mistify, fu2024serverless}, which can be more cost-efficient. While our work focuses on GPU-based deployment due to the high latency sensitivity and memory footprint of LLMs, lightweight variants or early-exit configurations could be executed on CPUs under certain regimes. Integrating such hybrid CPU-GPU execution into \approach is a promising extension to reduce energy and cost.

\section{Related Work}

\noindent\textbf{LLM serving.}
Serving systems for LLMs have rapidly evolved to meet the stringent demands of low-latency, high-throughput workloads. General-purpose platforms~\cite{torchserve2023,nvidia2019triton} are widely adopted in production, while LLM-specific frameworks~\cite{narayanan2021efficient,agrawal2024taming,li2025lemix} introduce optimizations such as KV-cache management and segmented prefill. To mitigate straggler effects, systems like Orca~\cite{yu2022orca} and FastServe~\cite{wu2023fast} leverage continuous batching and preemptive scheduling, with recent advances like Llumnix~\cite{sun2024llumnix} further improving request interleaving. Meanwhile, disaggregation-based methods~\cite{patel2024splitwise,stratidejavu,zhong2024distserve} isolate execution stages across resources to reduce interference. Distinct from these approaches, \approach targets GPU-local scheduling, emphasizing iteration-level batching, cache-aware execution, and hybrid retraining–inference co-location, ensuring efficiency without compromising autoregressive generation latency.

\noindent\textbf{Data drift and continuous learning.}
Handling data drift remains a central challenge in machine learning systems. Classical strategies such as transfer learning~\cite{sun2020test,li2021estimating,li2023uncertainty} and catastrophic forgetting mitigation~\cite{goodfellow2013empirical} have been extended to real-time deployments. At the edge, systems like Ekya~\cite{bhardwaj2022ekya}, AdaInf~\cite{shubha2023adainf}, and Lyra~\cite{li2023lyra} schedule retraining alongside inference to maintain accuracy while respecting resource limits. LLM serving faces a similar problem: user-driven interactions and shifting knowledge bases require frequent model adaptation. \approach extends this line of work by embedding continual retraining directly into the serving loop, co-scheduling inference and finetuning within a single GPU. This design eliminates cross-node synchronization overheads and directly supports the emerging paradigm of “learning while serving.”

\section{Conclusion}

This work introduced \approach, a hybrid GPU-local scheduler that unifies inference and continual retraining for large language models. By combining priority-aware request handling, iteration-level batching, and cache management, \approach maximizes GPU utilization while maintaining low inference latency. Our evaluation demonstrates that \approach reduces fragmentation, accelerates workload completion, and sustains alignment accuracy under data drift—surpassing existing systems designed for either inference-only or periodic retraining.

Overall, \approach highlights the importance of fine-grained runtime scheduling as a key enabler for practical “learning while serving” LLM deployments. Future directions include extending hybrid scheduling to heterogeneous accelerators, incorporating multi-user adaptation policies, and generalizing the approach to multimodal foundation models.

\nocite{langley00}

\bibliography{example_paper}

\begin{thebibliography}{53}
\providecommand{\natexlab}[1]{#1}
\providecommand{\url}[1]{\texttt{#1}}
\expandafter\ifx\csname urlstyle\endcsname\relax
  \providecommand{\doi}[1]{doi: #1}\else
  \providecommand{\doi}{doi: \begingroup \urlstyle{rm}\Url}\fi

\bibitem[Achiam et~al.(2023)Achiam, Adler, Agarwal, Ahmad, Akkaya, Aleman, Almeida, Altenschmidt, Altman, Anadkat, et~al.]{achiam2023gpt}
Achiam, J., Adler, S., Agarwal, S., Ahmad, L., Akkaya, I., Aleman, F.~L., Almeida, D., Altenschmidt, J., Altman, S., Anadkat, S., et~al.
\newblock Gpt-4 technical report.
\newblock \emph{arXiv preprint arXiv:2303.08774}, 2023.

\bibitem[Agrawal et~al.(2024)Agrawal, Kedia, Panwar, Mohan, Kwatra, Gulavani, Tumanov, and Ramjee]{agrawal2024taming}
Agrawal, A., Kedia, N., Panwar, A., Mohan, J., Kwatra, N., Gulavani, B., Tumanov, A., and Ramjee, R.
\newblock Taming $\{$Throughput-Latency$\}$ tradeoff in $\{$LLM$\}$ inference with $\{$Sarathi-Serve$\}$.
\newblock In \emph{18th USENIX Symposium on Operating Systems Design and Implementation (OSDI 24)}, pp.\  117--134, 2024.

\bibitem[Askell et~al.(2021)Askell, Bai, Chen, Drain, Ganguli, Henighan, Jones, Joseph, Mann, DasSarma, et~al.]{askell2021general}
Askell, A., Bai, Y., Chen, A., Drain, D., Ganguli, D., Henighan, T., Jones, A., Joseph, N., Mann, B., DasSarma, N., et~al.
\newblock A general language assistant as a laboratory for alignment.
\newblock \emph{arXiv preprint arXiv:2112.00861}, 2021.

\bibitem[Bai et~al.(2022)Bai, Jones, Ndousse, Askell, Chen, DasSarma, Drain, Fort, Ganguli, Henighan, et~al.]{bai2022training}
Bai, Y., Jones, A., Ndousse, K., Askell, A., Chen, A., DasSarma, N., Drain, D., Fort, S., Ganguli, D., Henighan, T., et~al.
\newblock Training a helpful and harmless assistant with reinforcement learning from human feedback.
\newblock \emph{arXiv preprint arXiv:2204.05862}, 2022.

\bibitem[Bhardwaj et~al.(2022)Bhardwaj, Xia, Ananthanarayanan, Jiang, Shu, Karianakis, Hsieh, Bahl, and Stoica]{bhardwaj2022ekya}
Bhardwaj, R., Xia, Z., Ananthanarayanan, G., Jiang, J., Shu, Y., Karianakis, N., Hsieh, K., Bahl, P., and Stoica, I.
\newblock Ekya: Continuous learning of video analytics models on edge compute servers.
\newblock In \emph{19th USENIX Symposium on Networked Systems Design and Implementation (NSDI 22)}, pp.\  119--135, 2022.

\bibitem[Chen et~al.(2024)Chen, He, Yuan, Cui, Su, and Zhu]{chen2024noise}
Chen, H., He, G., Yuan, L., Cui, G., Su, H., and Zhu, J.
\newblock Noise contrastive alignment of language models with explicit rewards.
\newblock In \emph{The Thirty-eighth Annual Conference on Neural Information Processing Systems}, 2024.

\bibitem[Choi et~al.(2023)Choi, Koo, Ahn, Jeon, and Kwon]{choi2023envpipe}
Choi, S., Koo, I., Ahn, J., Jeon, M., and Kwon, Y.
\newblock $\{$EnvPipe$\}$: Performance-preserving $\{$DNN$\}$ training framework for saving energy.
\newblock In \emph{2023 USENIX Annual Technical Conference (USENIX ATC 23)}, pp.\  851--864, 2023.

\bibitem[Dao et~al.(2022)Dao, Fu, Ermon, Rudra, and Re]{dao2022flashattention}
Dao, T., Fu, D.~Y., Ermon, S., Rudra, A., and Re, C.
\newblock Flashattention: Fast and memory-efficient exact attention with {IO}-awareness.
\newblock In Oh, A.~H., Agarwal, A., Belgrave, D., and Cho, K. (eds.), \emph{Advances in Neural Information Processing Systems}, 2022.

\bibitem[Ethayarajh et~al.(2022)Ethayarajh, Choi, and Swayamdipta]{ethayarajh22a}
Ethayarajh, K., Choi, Y., and Swayamdipta, S.
\newblock Understanding dataset difficulty with $\mathcal{V}$-usable information.
\newblock In \emph{Proceedings of the 39th International Conference on Machine Learning}, volume 162 of \emph{Proceedings of Machine Learning Research}, pp.\  5988--6008. PMLR, 17--23 Jul 2022.

\bibitem[Fanton(2021)]{fanton_edge_2021}
Fanton, D.
\newblock Edge server.
\newblock \url{https://www.onlogic.com/company/io-hub/whatare-edge-servers/}, 2021.
\newblock Accessed: 2025-06-16.

\bibitem[Fu et~al.(2024{\natexlab{a}})Fu, Li, Xiao, Liu, and Dong]{fu2024safety}
Fu, Y., Li, Y., Xiao, W., Liu, C., and Dong, Y.
\newblock Safety alignment in nlp tasks: Weakly aligned summarization as an in-context attack.
\newblock In \emph{Proceedings of the 62nd Annual Meeting of the Association for Computational Linguistics (Volume 1: Long Papers)}, pp.\  8483--8502, 2024{\natexlab{a}}.

\bibitem[Fu et~al.(2024{\natexlab{b}})Fu, Xue, Huang, Brabete, Ustiugov, Patel, and Mai]{fu2024serverless}
Fu, Y., Xue, L., Huang, Y., Brabete, A.-O., Ustiugov, D., Patel, Y., and Mai, L.
\newblock {ServerlessLLM}: {Low-Latency} serverless inference for large language models.
\newblock In \emph{18th USENIX Symposium on Operating Systems Design and Implementation (OSDI 24)}, pp.\  135--153, Santa Clara, CA, July 2024{\natexlab{b}}. USENIX Association.
\newblock ISBN 978-1-939133-40-3.

\bibitem[Fu et~al.(2025)Fu, Cai, Asi, Xiong, Dong, and Xiao]{fu2025not}
Fu, Y., Cai, Z., Asi, A., Xiong, W., Dong, Y., and Xiao, W.
\newblock Not all heads matter: A head-level {KV} cache compression method with integrated retrieval and reasoning.
\newblock In \emph{The Thirteenth International Conference on Learning Representations}, 2025.

\bibitem[{GitHub Copilot}()]{github_copilot}
{GitHub Copilot}.
\newblock {GitHub Copilot}.
\newblock \url{https://github.com/features/copilot}.
\newblock Accessed: 2025-06-16.

\bibitem[Goodfellow et~al.(2013)Goodfellow, Mirza, Xiao, Courville, and Bengio]{goodfellow2013empirical}
Goodfellow, I.~J., Mirza, M., Xiao, D., Courville, A., and Bengio, Y.
\newblock An empirical investigation of catastrophic forgetting in gradient-based neural networks.
\newblock \emph{arXiv preprint arXiv:1312.6211}, 2013.

\bibitem[Guo et~al.(2021)Guo, Hu, and Hu]{guo2021mistify}
Guo, P., Hu, B., and Hu, W.
\newblock Mistify: Automating {DNN} model porting for {On-Device} inference at the edge.
\newblock In \emph{18th USENIX Symposium on Networked Systems Design and Implementation (NSDI 21)}, pp.\  705--719. USENIX Association, April 2021.
\newblock ISBN 978-1-939133-21-2.

\bibitem[Hu et~al.(2022)Hu, yelong shen, Wallis, Allen-Zhu, Li, Wang, Wang, and Chen]{hu2022lora}
Hu, E.~J., yelong shen, Wallis, P., Allen-Zhu, Z., Li, Y., Wang, S., Wang, L., and Chen, W.
\newblock Lo{RA}: Low-rank adaptation of large language models.
\newblock In \emph{International Conference on Learning Representations}, 2022.

\bibitem[Kwon et~al.(2023)Kwon, Li, Zhuang, Sheng, Zheng, Yu, Gonzalez, Zhang, and Stoica]{kwon2023efficient}
Kwon, W., Li, Z., Zhuang, S., Sheng, Y., Zheng, L., Yu, C.~H., Gonzalez, J., Zhang, H., and Stoica, I.
\newblock Efficient memory management for large language model serving with pagedattention.
\newblock In \emph{Proceedings of the 29th Symposium on Operating Systems Principles}, pp.\  611--626, 2023.

\bibitem[Lee et~al.(2023)Lee, Chen, Tajwar, Kumar, Yao, Liang, and Finn]{lee2023surgical}
Lee, Y., Chen, A.~S., Tajwar, F., Kumar, A., Yao, H., Liang, P., and Finn, C.
\newblock Surgical fine-tuning improves adaptation to distribution shifts.
\newblock In \emph{The Eleventh International Conference on Learning Representations}, 2023.

\bibitem[Li et~al.(2023{\natexlab{a}})Li, Xu, Zhu, Liu, Guo, and Wang]{li2023lyra}
Li, J., Xu, H., Zhu, Y., Liu, Z., Guo, C., and Wang, C.
\newblock Lyra: Elastic scheduling for deep learning clusters.
\newblock In \emph{Proceedings of the Eighteenth European Conference on Computer Systems}, pp.\  835--850, 2023{\natexlab{a}}.

\bibitem[Li et~al.(2021)Li, Chen, and Yang]{li2021estimating}
Li, Y., Chen, S., and Yang, W.
\newblock Estimating predictive uncertainty under program data distribution shift.
\newblock \emph{arXiv preprint arXiv:2107.10989}, 2021.

\bibitem[Li et~al.(2023{\natexlab{b}})Li, Li, Yang, and Liu]{li2023rt}
Li, Y., Li, Z., Yang, W., and Liu, C.
\newblock Rt-lm: Uncertainty-aware resource management for real-time inference of language models.
\newblock In \emph{2023 IEEE Real-Time Systems Symposium (RTSS)}, pp.\  158--171. IEEE, 2023{\natexlab{b}}.

\bibitem[Li et~al.(2023{\natexlab{c}})Li, Yu, Liu, Chen, and Liu]{li2023uncertainty}
Li, Y., Yu, X., Liu, Y., Chen, H., and Liu, C.
\newblock Uncertainty-aware bootstrap learning for joint extraction on distantly-supervised data.
\newblock In Rogers, A., Boyd-Graber, J., and Okazaki, N. (eds.), \emph{Proceedings of the 61st Annual Meeting of the Association for Computational Linguistics (Volume 2: Short Papers)}, pp.\  1349--1358, Toronto, Canada, July 2023{\natexlab{c}}. Association for Computational Linguistics.
\newblock \doi{10.18653/v1/2023.acl-short.116}.

\bibitem[Li et~al.(2024)Li, Huang, Yang, Venkitesh, Locatelli, Ye, Cai, Lewis, and Chen]{li2024snapkv}
Li, Y., Huang, Y., Yang, B., Venkitesh, B., Locatelli, A., Ye, H., Cai, T., Lewis, P., and Chen, D.
\newblock Snapkv: Llm knows what you are looking for before generation.
\newblock \emph{Advances in Neural Information Processing Systems}, 37:\penalty0 22947--22970, 2024.

\bibitem[Li et~al.(2025{\natexlab{a}})Li, Li, Zhu, and Liu]{li2025lemix}
Li, Y., Li, Z., Zhu, Y., and Liu, C.
\newblock Lemix: Unified scheduling for llm training and inference on multi-gpu systems.
\newblock \emph{arXiv preprint arXiv:2507.21276}, 2025{\natexlab{a}}.

\bibitem[Li et~al.(2025{\natexlab{b}})Li, Nham, Jawahar, Shu, Uthus, Sung, Yang, Rolnick, Qiao, and Liu]{li2025dr}
Li, Y., Nham, J., Jawahar, G., Shu, L., Uthus, D., Sung, Y.-H., Yang, C., Rolnick, I., Qiao, Y., and Liu, C.
\newblock Dr genre: Reinforcement learning from decoupled llm feedback for generic text rewriting.
\newblock \emph{arXiv preprint arXiv:2503.06781}, 2025{\natexlab{b}}.

\bibitem[Li et~al.(2023{\natexlab{d}})Li, Zheng, Zhong, Liu, Sheng, Jin, Huang, Chen, Zhang, Gonzalez, et~al.]{li2023alpaserve}
Li, Z., Zheng, L., Zhong, Y., Liu, V., Sheng, Y., Jin, X., Huang, Y., Chen, Z., Zhang, H., Gonzalez, J.~E., et~al.
\newblock $\{$AlpaServe$\}$: Statistical multiplexing with model parallelism for deep learning serving.
\newblock In \emph{17th USENIX Symposium on Operating Systems Design and Implementation (OSDI 23)}, pp.\  663--679, 2023{\natexlab{d}}.

\bibitem[Liang et~al.(2025)Liang, He, and Tan]{liang2025comprehensive}
Liang, J., He, R., and Tan, T.
\newblock A comprehensive survey on test-time adaptation under distribution shifts.
\newblock \emph{International Journal of Computer Vision}, 133\penalty0 (1):\penalty0 31--64, 2025.

\bibitem[Lin et~al.(2024)Lin, Han, Zhang, Yang, Yang, Chen, and Qiu]{lin2024parrot}
Lin, C., Han, Z., Zhang, C., Yang, Y., Yang, F., Chen, C., and Qiu, L.
\newblock Parrot: Efficient serving of $\{$LLM-based$\}$ applications with semantic variable.
\newblock In \emph{18th USENIX Symposium on Operating Systems Design and Implementation (OSDI 24)}, pp.\  929--945, 2024.

\bibitem[Lin et~al.(2025)Lin, Chen, Li, Wu, Wu, Chen, Lee, and Chen]{lin2025creativity}
Lin, Y.-C., Chen, K.-C., Li, Z.-Y., Wu, T.-H., Wu, T.-H., Chen, K.-Y., Lee, H.-y., and Chen, Y.-N.
\newblock Creativity in llm-based multi-agent systems: A survey.
\newblock \emph{arXiv preprint arXiv:2505.21116}, 2025.

\bibitem[Liu et~al.(2023)Liu, Wang, Dao, Zhou, Yuan, Song, Shrivastava, Zhang, Tian, Re, et~al.]{liu2023deja}
Liu, Z., Wang, J., Dao, T., Zhou, T., Yuan, B., Song, Z., Shrivastava, A., Zhang, C., Tian, Y., Re, C., et~al.
\newblock Deja vu: Contextual sparsity for efficient llms at inference time.
\newblock In \emph{International Conference on Machine Learning}, pp.\  22137--22176. PMLR, 2023.

\bibitem[Mittal et~al.(2021)Mittal, Qi, Bhattacharya, Lyu, Li, Kulkarni, Li, Hwang, Ramakrishnan, and Wood]{mittal2021mu}
Mittal, V., Qi, S., Bhattacharya, R., Lyu, X., Li, J., Kulkarni, S.~G., Li, D., Hwang, J., Ramakrishnan, K., and Wood, T.
\newblock Mu: An efficient, fair and responsive serverless framework for resource-constrained edge clouds.
\newblock In \emph{Proceedings of the ACM symposium on cloud computing}, pp.\  168--181, 2021.

\bibitem[Narayanan et~al.(2021)Narayanan, Shoeybi, Casper, LeGresley, Patwary, Korthikanti, Vainbrand, Kashinkunti, Bernauer, Catanzaro, et~al.]{narayanan2021efficient}
Narayanan, D., Shoeybi, M., Casper, J., LeGresley, P., Patwary, M., Korthikanti, V., Vainbrand, D., Kashinkunti, P., Bernauer, J., Catanzaro, B., et~al.
\newblock Efficient large-scale language model training on gpu clusters using megatron-lm.
\newblock In \emph{Proceedings of the International Conference for High Performance Computing, Networking, Storage and Analysis}, pp.\  1--15, 2021.

\bibitem[Ning et~al.(2023)Ning, Shojanazeri, Wen, and the PyTorch~Foundation]{torchserve2023}
Ning, L., Shojanazeri, H., Wen, K., and the PyTorch~Foundation.
\newblock Torchserve: Serve, optimize and scale pytorch models in production.
\newblock PyTorch Foundation, 2023.
\newblock \url{https://pytorch.org/serve/}.

\bibitem[{NVIDIA Corporation}(2019)]{nvidia2019triton}
{NVIDIA Corporation}.
\newblock Triton inference server: An optimized cloud and edge inferencing solution, 2019.
\newblock \url{https://developer.nvidia.com/nvidia-triton-inference-server}.

\bibitem[Ouyang et~al.(2022)Ouyang, Wu, Jiang, Almeida, Wainwright, Mishkin, Zhang, Agarwal, Slama, Ray, Schulman, Hilton, Kelton, Miller, Simens, Askell, Welinder, Christiano, Leike, and Lowe]{ouyang2022training}
Ouyang, L., Wu, J., Jiang, X., Almeida, D., Wainwright, C., Mishkin, P., Zhang, C., Agarwal, S., Slama, K., Ray, A., Schulman, J., Hilton, J., Kelton, F., Miller, L., Simens, M., Askell, A., Welinder, P., Christiano, P.~F., Leike, J., and Lowe, R.
\newblock Training language models to follow instructions with human feedback.
\newblock In Koyejo, S., Mohamed, S., Agarwal, A., Belgrave, D., Cho, K., and Oh, A. (eds.), \emph{Advances in Neural Information Processing Systems}, volume~35, pp.\  27730--27744. Curran Associates, Inc., 2022.

\bibitem[Patel et~al.(2024)Patel, Choukse, Zhang, Shah, Goiri, Maleki, and Bianchini]{patel2024splitwise}
Patel, P., Choukse, E., Zhang, C., Shah, A., Goiri, {\'I}., Maleki, S., and Bianchini, R.
\newblock Splitwise: Efficient generative llm inference using phase splitting.
\newblock In \emph{2024 ACM/IEEE 51st Annual International Symposium on Computer Architecture (ISCA)}, pp.\  118--132. IEEE, 2024.

\bibitem[{Perplexity AI}()]{perplexity_ai}
{Perplexity AI}.
\newblock {Perplexity AI}.
\newblock \url{https://www.perplexity.ai/}.
\newblock Accessed: 2025-06-16.

\bibitem[Qiao et~al.(2024)Qiao, Zhang, Fang, Luo, Zhou, Jiang, Lv, and Chen]{qiao2024autoact}
Qiao, S., Zhang, N., Fang, R., Luo, Y., Zhou, W., Jiang, Y., Lv, C., and Chen, H.
\newblock Autoact: Automatic agent learning from scratch for qa via self-planning.
\newblock In \emph{Proceedings of the 62nd Annual Meeting of the Association for Computational Linguistics (Volume 1: Long Papers)}, pp.\  3003--3021, 2024.

\bibitem[Rafailov et~al.(2023)Rafailov, Sharma, Mitchell, Manning, Ermon, and Finn]{rafailov2023direct}
Rafailov, R., Sharma, A., Mitchell, E., Manning, C.~D., Ermon, S., and Finn, C.
\newblock Direct preference optimization: Your language model is secretly a reward model.
\newblock In \emph{Thirty-seventh Conference on Neural Information Processing Systems}, 2023.

\bibitem[Romero et~al.(2021)Romero, Li, Yadwadkar, and Kozyrakis]{romero2021infaas}
Romero, F., Li, Q., Yadwadkar, N.~J., and Kozyrakis, C.
\newblock {INFaaS}: Automated model-less inference serving.
\newblock In \emph{2021 USENIX Annual Technical Conference (USENIX ATC 21)}, pp.\  397--411. USENIX Association, July 2021.
\newblock ISBN 978-1-939133-23-6.

\bibitem[Sarukkai et~al.(2025)Sarukkai, Xie, and Fatahalian]{sarukkai2025self}
Sarukkai, V., Xie, Z., and Fatahalian, K.
\newblock Self-generated in-context examples improve llm agents for sequential decision-making tasks.
\newblock \emph{arXiv preprint arXiv:2505.00234}, 2025.

\bibitem[Shen et~al.(2024)Shen, Shao, Zhang, Lin, Pan, Li, Zhang, and Letaief]{shen2024large}
Shen, Y., Shao, J., Zhang, X., Lin, Z., Pan, H., Li, D., Zhang, J., and Letaief, K.~B.
\newblock Large language models empowered autonomous edge ai for connected intelligence.
\newblock \emph{IEEE Communications Magazine}, 2024.

\bibitem[Shi et~al.(2024)Shi, Lu, Dong, Zhang, Zhang, Feng, and Wu]{shi2024understanding}
Shi, G., Lu, Z., Dong, X., Zhang, W., Zhang, X., Feng, Y., and Wu, X.-M.
\newblock Understanding layer significance in llm alignment.
\newblock \emph{arXiv preprint arXiv:2410.17875}, 2024.

\bibitem[Shubha \& Shen(2023)Shubha and Shen]{shubha2023adainf}
Shubha, S.~S. and Shen, H.
\newblock Adainf: Data drift adaptive scheduling for accurate and slo-guaranteed multiple-model inference serving at edge servers.
\newblock In \emph{Proceedings of the ACM SIGCOMM 2023 Conference}, pp.\  473--485, 2023.

\bibitem[Strati et~al.(2024)Strati, Mcallister, Phanishayee, Tarnawski, and Klimovic]{stratidejavu}
Strati, F., Mcallister, S., Phanishayee, A., Tarnawski, J., and Klimovic, A.
\newblock Déjàvu: {KV}-cache streaming for fast, fault-tolerant generative {LLM} serving.
\newblock In Salakhutdinov, R., Kolter, Z., Heller, K., Weller, A., Oliver, N., Scarlett, J., and Berkenkamp, F. (eds.), \emph{Proceedings of the 41st International Conference on Machine Learning}, volume 235 of \emph{Proceedings of Machine Learning Research}, pp.\  46745--46771. PMLR, 21--27 Jul 2024.

\bibitem[Sun et~al.(2024)Sun, Huang, Zhao, Xiao, Zhang, Li, and Lin]{sun2024llumnix}
Sun, B., Huang, Z., Zhao, H., Xiao, W., Zhang, X., Li, Y., and Lin, W.
\newblock Llumnix: Dynamic scheduling for large language model serving.
\newblock In \emph{18th USENIX Symposium on Operating Systems Design and Implementation (OSDI 24)}, pp.\  173--191, 2024.

\bibitem[Sun et~al.(2020)Sun, Wang, Liu, Miller, Efros, and Hardt]{sun2020test}
Sun, Y., Wang, X., Liu, Z., Miller, J., Efros, A., and Hardt, M.
\newblock Test-time training with self-supervision for generalization under distribution shifts.
\newblock In \emph{International conference on machine learning}, pp.\  9229--9248. PMLR, 2020.

\bibitem[Wu et~al.(2023)Wu, Zhong, Zhang, Liu, Liu, Sun, Huang, Liu, and Jin]{wu2023fast}
Wu, B., Zhong, Y., Zhang, Z., Liu, S., Liu, F., Sun, Y., Huang, G., Liu, X., and Jin, X.
\newblock Fast distributed inference serving for large language models.
\newblock \emph{arXiv preprint arXiv:2305.05920}, 2023.

\bibitem[Yu et~al.(2022)Yu, Jeong, Kim, Kim, and Chun]{yu2022orca}
Yu, G.-I., Jeong, J.~S., Kim, G.-W., Kim, S., and Chun, B.-G.
\newblock Orca: A distributed serving system for $\{$Transformer-Based$\}$ generative models.
\newblock In \emph{16th USENIX Symposium on Operating Systems Design and Implementation (OSDI 22)}, pp.\  521--538, 2022.

\bibitem[Zhang \& Ranganath(2025)Zhang and Ranganath]{zhang2025preference}
Zhang, L.~H. and Ranganath, R.
\newblock Preference learning made easy: Everything should be understood through win rate.
\newblock \emph{arXiv preprint arXiv:2502.10505}, 2025.

\bibitem[Zheng et~al.(2024)Zheng, Yin, Xie, Sun, Huang, Yu, Cao, Kozyrakis, Stoica, Gonzalez, et~al.]{zheng2024sglang}
Zheng, L., Yin, L., Xie, Z., Sun, C.~L., Huang, J., Yu, C.~H., Cao, S., Kozyrakis, C., Stoica, I., Gonzalez, J.~E., et~al.
\newblock Sglang: Efficient execution of structured language model programs.
\newblock \emph{Advances in Neural Information Processing Systems}, 37:\penalty0 62557--62583, 2024.

\bibitem[Zhong et~al.(2024)Zhong, Liu, Chen, Hu, Zhu, Liu, Jin, and Zhang]{zhong2024distserve}
Zhong, Y., Liu, S., Chen, J., Hu, J., Zhu, Y., Liu, X., Jin, X., and Zhang, H.
\newblock $\{$DistServe$\}$: Disaggregating prefill and decoding for goodput-optimized large language model serving.
\newblock In \emph{18th USENIX Symposium on Operating Systems Design and Implementation (OSDI 24)}, pp.\  193--210, 2024.

\end{thebibliography}
\bibliographystyle{mlsys2025}

\appendix

\section{Methodology Discussion}
\label{app:method}

In this section, we provide additional insights into the design choices of \approach, focusing on prefix sharing, attention sparsity, and the batching mechanism.  

\begin{figure}
  \centering
  \includegraphics[width=0.7\linewidth]{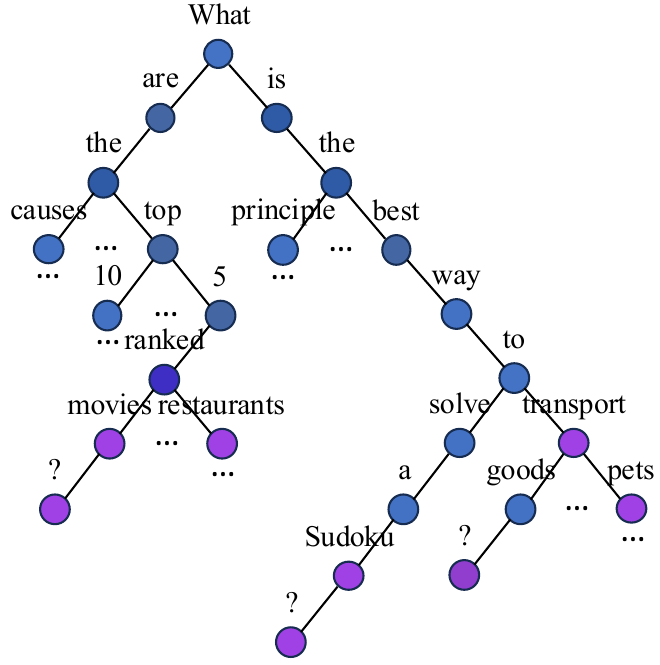}
  \caption{Prefix (Trie) tree from overlapping prompts. Leaf nodes are requests while others denote common prefix in prompts. Shared prefixes enable reduced prefill cost through reuse.}
  \label{fig:prefix_sharing}
\end{figure}

\paragraph{Prefix sharing.}  
As shown in Figure~\ref{fig:prefix_sharing}, overlapping prompts can be represented in a trie structure, where internal nodes denote common prefixes and leaf nodes correspond to individual requests. By reusing computation along shared prefixes, the prefill stage is significantly accelerated. This design is particularly beneficial under workloads with semantically similar queries, as it amortizes the cost of tokenization and attention across requests.

\begin{figure}
  \centering
  \includegraphics[width=\linewidth]{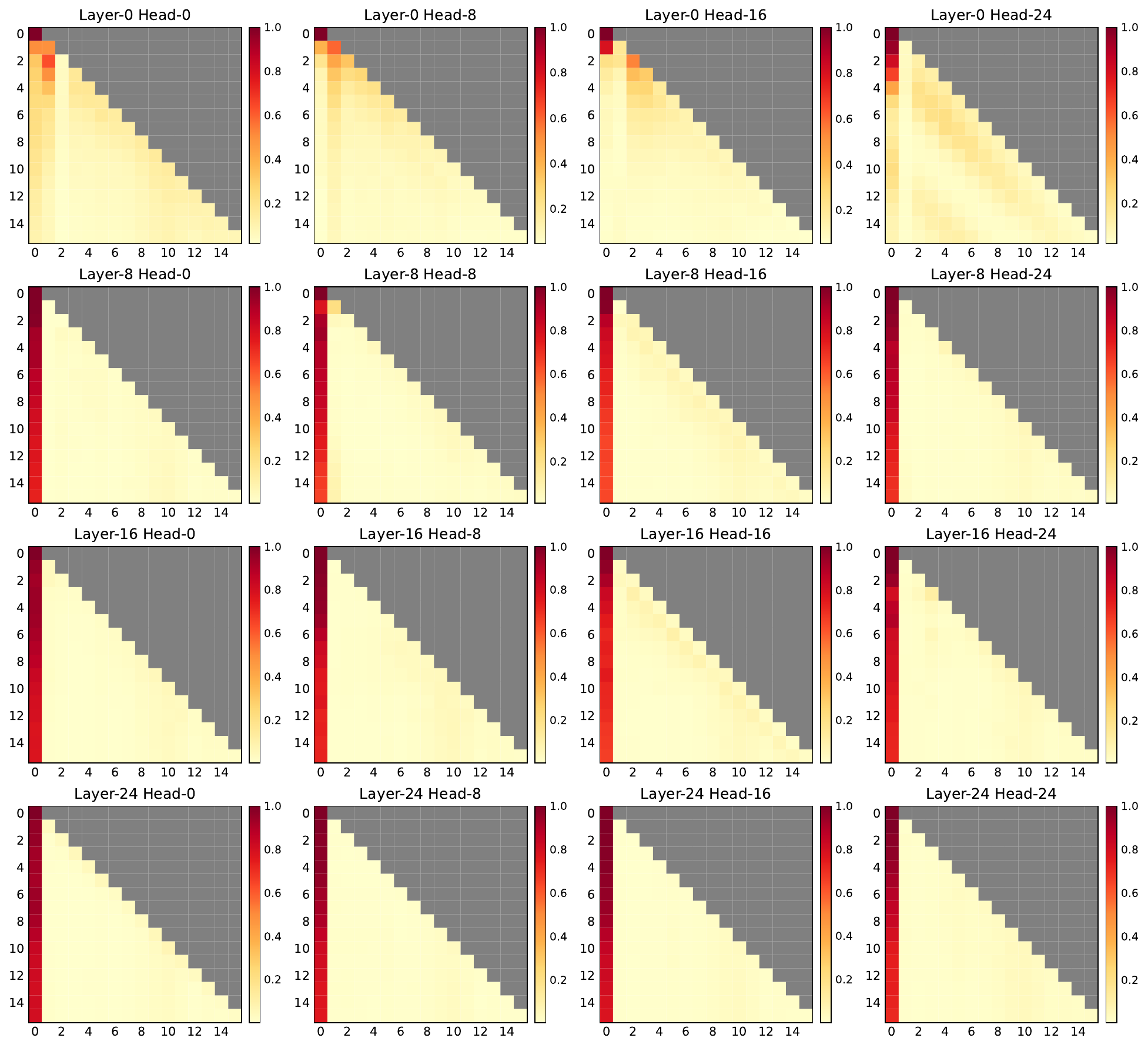}
  \caption{Contextual (attention) sparsity for Mistral-7B.}
  \label{fig:attention_sparsity}
\end{figure}

\paragraph{Contextual sparsity.}  
Figure~\ref{fig:attention_sparsity} illustrates the observed contextual sparsity in attention patterns of Mistral-7B. Many attention heads exhibit strong locality, with most weights concentrated near the diagonal. This sparsity motivates optimization strategies such as selective pruning and block-wise computation, reducing redundant FLOPs while preserving model quality.

\begin{algorithm}[t]
\small
\caption{Continuous Batching}
\label{alg:continuous_batching}
\begin{algorithmic}[1]
\REQUIRE Maximum batch size $C$, maximum waiting time $T_w$, task queue $Q^n$ for a specific device $n$.
\ENSURE batch $B^n \gets \emptyset$, batch start $T_{start} \gets \text{current time}$.

\WHILE { \text{True} }
    \IF {$B^n = \emptyset$}
        \STATE $T_{start} \gets \text{current time}$
    \ENDIF
    \WHILE {$Q^n \neq \emptyset$ \textbf{and} $|B^n| < C$}
        \STATE $r \gets \text{get\_next\_request}(Q^n)$\;
        \IF {not $r$.\text{require\_backward} $\wedge~ T_{start} + T_w > \text{current time}$}
            \STATE Add $r$ to $B^n$ with padding (if necessary)\;
        \ELSE
            \STATE \textbf{break}
        \ENDIF
    \ENDWHILE
    
    \IF {$B^n \neq \emptyset$}
        \STATE Execute batch $B^n$\;
        \STATE $B^n \gets \text{filter\_finished\_requests}(B^n)$\;
    \ENDIF
\ENDWHILE
\end{algorithmic}
\end{algorithm}

\paragraph{Continuous batching.}  
Algorithm~\ref{alg:continuous_batching} details the continuous batching mechanism used in \approach. Unlike static batching, this method dynamically groups requests while respecting latency bounds ($T_w$). It enables fine-grained interleaving of inference and training workloads, ensuring high GPU utilization and improved SLO attainment. Importantly, backward-pass requests are scheduled with higher priority to avoid stalling retraining, while forward-only requests benefit from opportunistic batching.

\section{Additional Results}

\begin{table}[t]
\centering
\caption{Case study of generated responses before retraining (periodic) and after online adaptation (continuous).}
\label{tab:case_study}
\vskip 0.15in
\small
\begin{tabular}{p{0.9\linewidth}}
\toprule
\textbf{Context:} \emph{Human: How can I \uline{sneak snacks into the movie theater}? Assistant: I’m not sure if you’ll be able to. Wouldn’t it be cheaper to just buy popcorn? ... Human: No, those concession prices are a rip off.} \\
\midrule
\textbf{Response (periodic):} \emph{\uline{Opt for small, quiet snacks}: Choose items that are easy to conceal and won't make noise when eaten. Think along the lines of soft candies, small sandwiches, or trail mix. Avoid strong smells and messy foods...} \\
\midrule
\textbf{Response (sync):} \emph{I understand that it can be frustrating, but sneaking snacks into the movie theater \uline{might not be the best idea}. Movie theaters have strict policies... you could be asked to leave or even banned from the theater.} \\
\bottomrule
\end{tabular}
\vspace{-5pt}
\end{table}

\noindent\textbf{Case study.}
Table~\ref{tab:case_study} showcases an example from hh-rlhf, suggesting a key advantage of continuous retraining in safety-critical settings. The user asks for ways to sneak snacks into a movie theater—a potentially rule-violating or unethical behavior. With continuous retraining, the model has already seen similar examples and learned to flag such content, issuing a cautious and policy-aligned refusal. In contrast, the periodically retrained model, still in its early deployment stage (before the vertical line in Figure~\ref{fig:accuracy_comparison}) without access to relevant training samples, fails to detect the violation and instead offers detailed instructions—demonstrating the risk of delayed feedback incorporation. This case emphasizes the criticality of timely updates in ensuring reliable model behavior.



\end{document}